\documentclass[letterpaper, journal]{IEEEtran}  
\IEEEoverridecommandlockouts                              

\def\endthebibliography{%
  \def\@noitemerr{\@latex@warning{Empty `thebibliography' environment}}%
  \endlist
}

\usepackage{cite}
\usepackage{graphicx} 
\usepackage[utf8]{inputenc}
\usepackage{amsmath} 
\usepackage{amssymb}  
\usepackage{fixmath}  
\usepackage{array}  
\usepackage[usenames,dvipsnames]{xcolor}  
\usepackage{hyperref}
\hypersetup{
    colorlinks,
    linkcolor = BlueViolet,
    urlcolor  = Blue,
    citecolor = Black,
    anchorcolor = Black
}
\usepackage[nohyperlinks,printonlyused,nolist]{acronym}
\usepackage{subcaption}
\usepackage{multirow}

\newtheorem{definition}{Definition}

\usepackage{algorithm}
\usepackage{algpseudocode}

\usepackage{tikz}
\usepackage{pgfplots}
\pgfplotsset{compat=newest}
\usepgfplotslibrary{groupplots}

\begin{acronym}[SICNav]
    \acrodef{EKF}{Extended Kalman Filter}
    \acrodef{KF}{Kalman Filter}
    \acrodefplural{KFs}{Kalman Filters}
    \acro{MPC}{Model Predictive Control}
    \acro{CVAE}{Conditional Variational Autoencoder}
    \acro{DWA}{Dynamic Window Approach}
    \acro{EB}{Elastic Band}
    \acro{FRP}{Freezing Robot Problem}
    \acro{GP}{Gaussian Process}
    \acrodefplural{GP}{Gaussian Processes}
    \acro{IGP}{Interacting Gaussian processes}
    \acro{IL}{Imitation Learning}
    \acro{IRL}{Inverse Reinforcement Learning}
    \acro{LSTM}{Long-Short Term Memory}
    \acrodefplural{LSTM}{Long-Short Term Memory networks}
    \acro{MAP}{Maximum a Posteriori}
    \acro{MLP}{Multi-Layer Perceptron}
    \acro{MPC}{Model Predictive Control}
    \acro{ORCA}{Online Reciprocal Collision Avoidance}
    \acro{PCLRHC}{Partially Closed Loop Receding Horizon Control}
    \acro{RHC}{Receding Horizon Control}
    \acro{RL}{Reinforcement Learning}
    \acro{ROS}{Robot Operating System}
    \acro{SARL}{Socially Attentive Reinforcement Learning}
    \acro{RGL}{Relational Graph Learning}
    \acro{SOGM}{Spatiotemporal Occupancy Grid Map}
    \acro{TEB}{Timed Elastic Band}
    \acro{VAE}{Variational Autoencoder}
    \acro{CVMM}{Constant Velocity Motion Model}
    \acro{MATS}{Mixtures of Affine Time-varying Systems}
    \acro{AV}{Autonomous vehicle}
    \acrodefplural{AVs}{Autonomous vehicles}

    \acrodef{LP}{Linear Program}
    \acrodef{QCQP}{Quadratically Constrained Quadratic Program}
    \acrodef{MPCC}{Mathematical Program with Complementarity Constraints}
    \acrodef{KKT}{Karush-Kuhn-Tucker}
    \acrodef{LICQ}{Linear Independence Constraint Qualification}
    \acrodef{CRCQ}{Constant Rank Constraint Qualification}
    \acrodef{SCQ}{Slater's Constraint Qualification}

    \acrodef{OGM}{Occupancy Grid Map}
    \acrodef{NF}{Navigation Function}
    \acrodef{RVO}{Reciprocal Velocity Obstacle}
    \acrodefplural{RVO}{Reciprocal Velocity Obstacles}
    \acrodef{SICNav}{Safe and Interactive Crowd Navigation}

    \acrodef{MPC}{Model Predictive Control}
    \acrodef{CAMPC}{Collision Avoiding Model Predictive Control}
    \acrodef{RHC}{Receding Horizon Control}
    \acrodef{CLF}{control-Lyapunov Function}
    \acrodef{DWA}{Dynamic Window Approach}
    \acrodef{SFM}{Social Force Model}
    \acrodef{ESFM}{Extended Social Force Model}

    \acrodef{ORCA}{Optimal Reciprocal Collision Avoidance}
    \acrodef{VO}{Velocity Obstacle}
    \acrodef{CA}{Collision Avoiding}

    \acrodef{ADE}{Average Displacement Error}
    \acrodef{FDE}{Final Displacement Error}
    \acrodef{KDE}{Kernel Density Estimate}
    \acrodefplural{KDE}{Kernel Density Estimates}
    \acrodef{NLL}{Negative Log Likelihood}

    \acrodef{RMSE}{Root Mean Squared Error}
\end{acronym}

\def\MYTITLE{SICNav: Safe and Interactive Crowd Navigation using Model Predictive Control and Bilevel Optimization}
\title{\MYTITLE}

\author{Sepehr Samavi, James R. Han, Florian Shkurti, Angela P. Schoellig
\thanks{Sepehr Samavi, James R. Han and Florian Shkurti are with the University of Toronto Robotics Institute and the Vector Institute for Artificial Intelligence, Toronto, Canada. Angela P. Schoellig is with the Technical University of Munich, University of Toronto, the Vector Institute for Artificial Intelligence, and the Munich Institute for Robotics and Machine Intelligence (MIRMI), Munich, Germany. (emails: sepehr@robotics.utias.utoronto.ca, jamesr.han@mail.utoronto.ca, florian@cs.toronto.edu, angela.schoellig@tum.de)}
}

\newcommand{\set}[1]{\bigl\{#1\bigr\}}
\newcommand{\tuple}[1]{\left(#1\right)}
\newcommand{\st}{:} 

\newcommand{\norm}[1]{\bigl\lVert#1\bigr\rVert}
\newcommand{\pnorm}[2]{\norm{#2}_{#1}}
\newcommand{\twonorm}[1]{\pnorm{2}{#1}}

\newcommand{\Reals}{\mathbb R}

\renewcommand{\v}{\mathbf{v}}
\newcommand{\x}{\mathbf{x}}
\renewcommand{\u}{\mathbf{u}}

\newcommand{\f}{\mathbf{f}}
\newcommand{\h}{\mathbf{h}}

\newcommand{\Q}{\mathbf{Q}}
\newcommand{\R}{\mathbf{R}}
\renewcommand{\P}{\mathbf{P}}
\newcommand{\p}{\mathbf{p}}

\newcommand{\trans}[1]{#1^\top}

\usepackage{booktabs}

\newcommand*\centremathcell[1]{\omit\hfil$\displaystyle#1$\hfil\ignorespaces}

\DeclareMathOperator*{\argmax}{arg\,max}
\DeclareMathOperator*{\argmin}{arg\,min}
\DeclareMathOperator*{\minimize}{minimize}
\DeclareMathOperator*{\subjectto}{subject\ to}

\newcommand{\bbm}{\begin{bmatrix}}
\newcommand{\ebm}{\end{bmatrix}}
\DeclareMathAlphabet{\mbf}{OT1}{ptm}{b}{n}

\newcommand{\state}{\x}
\newcommand{\error}{\bar{\state}}
\newcommand{\init}{0}
\newcommand{\stateinit}{{\state_{\init}}}
\newcommand{\sspace}{\Reals^{4+4\numhumans}}
\newcommand{\control}{\u}

\newcommand{\cspace}{\Reals^2}
\newcommand{\action}{\control}

\newcommand{\at}[2]{{#1}_{#2}}
\newcommand{\kpone}{{t+1}}
\newcommand{\kmone}{{t-1}}
\newcommand{\stateat}[1]{\at{\state}{#1}}
\newcommand{\actionat}[1]{\at{\action}{#1}}
\newcommand{\controlat}[1]{\at{\control}{#1}}

\newcommand{\velvec}{\v}
\newcommand{\vel}{\velvec}

\newcommand{\lvel}{{v}}
\newcommand{\dist}{{d}}

\newcommand{\robidmarker}{r}
\newcommand{\idRob}{\robidmarker}

\newcommand{\id}[2]{{#2^{(#1)}}}
\newcommand{\rob}[1]{\id{\robidmarker}{#1}}
\newcommand{\goal}{\mathbf{g}}
\newcommand{\pref}{\text{pref}}

\newcommand{\robdyn}{\f}
\newcommand{\humdyn}{\h}

\newcommand{\numhumans}{N}
\newcommand{\numstatobs}{M}
\newcommand{\forallhumanslongset}{\mathbb{I}_{1:\numhumans}=\{1,\dots,\numhumans\}}
\newcommand{\forallhumans}{\forall \idA \in \mathbb{I}_{1:\numhumans}}
\newcommand{\forallstatobslongset}{\set{\numhumans+1,\dots,\numhumans+\numstatobs}}
\newcommand{\forallstatobs}{\forall \idStat \in \forallstatobslongset}

\newcommand{\actionhum}{\tilde{\control}}

\newcommand{\idA}{j}
\newcommand{\idB}{l}
\newcommand{\idStat}{\tilde{l}}
\newcommand{\colTau}{\tau}

\newcommand{\orcalinedir}{\mathbf{n}}

\newcommand{\orcaconstgeq}{\rho}

\newcommand{\orcasolnset}{\mathcal{O}}
\newcommand{\orcarlxsolnset}{\mathcal{O}_{rlx}}

\newcommand{\horiz}{T}

\newcommand{\stagecostsymb}{l}
\newcommand{\stagecost}[1]{\stagecostsymb(#1)}
\newcommand{\valfuncsymb}{V}
\newcommand{\termpenalsymb}{\valfuncsymb_{f}}
\newcommand{\termpenal}[1]{\termpenalsymb(#1)}

\newcommand{\forallactidcs}{\forall t\in\mathbb{I}_{0:\horiz-1}}

\newcommand{\slackvar}{\zeta}
\newcommand{\penal}{M}

\newcommand{\lagmul}{\lambda}
\newcommand{\lagmuls}{\mathbold{\lagmul}}
\newcommand{\lagfun}{\mathcal{L}}
\newcommand{\forallorcas}{\forall \idB \in \set{r,1,\dots,\numhumans} \setminus \set{j}}

\newcommand{\forallstatorcas}{\forallstatobs}

\newcommand{\varsoneorca}{{\vel}, {\slackvar}}
\newcommand{\varoneorca}{{\mathbold{\nu}}}
\newcommand{\varsoneorcastacked}{({\vel}, {\slackvar})}

\newcommand{\varoneorcalocopt}{{\bar{\mathbold{\nu}}}}
\newcommand{\varandparamoneorca}{\tuple{\varoneorca; \at{{\state}}{t}}}
\newcommand{\varandparamoneorcalocopt}{\tuple{\varoneorcalocopt; \at{\bar{\state}}{t}}}

\newcommand{\ithconstorcarel}[1]{\id{\idA}{\at{{c}}{#1}}}

\newcommand{\ithconstorcarelatbar}[1]{\ithconstorcarel{#1} \varandparamoneorcalocopt}
\newcommand{\gradorcavars}{\nabla_{\varoneorca}}

\newcommand{\bilevellocoptoneorca}{\tuple{\at{\bar{\state}}{t}, \varoneorcalocopt}}
\usepackage{balance}

\definecolor{blue}{rgb}{0,0,0}

\begin{document}
\maketitle

\begin{abstract}
Robots need to predict and react to human motions to navigate through a crowd without collisions. Many existing methods decouple prediction from planning, which does not account for the interaction between robot and human motions and can lead to the robot getting stuck. We propose SICNav, a \ac{MPC} method that \textit{jointly} solves for robot motion and predicted crowd motion in closed-loop. We model each human in the crowd to be following an \ac{ORCA} scheme and embed that model as a constraint in the robot's local planner, resulting in a bilevel nonlinear \ac{MPC} optimization problem. We use a KKT-reformulation to cast the bilevel problem as a single level and use a nonlinear solver to optimize. Our \ac{MPC} method can influence pedestrian motion while explicitly satisfying safety constraints in a single-robot multi-human environment. We analyze the performance of SICNav in {\color{blue} two} simulation environments {\color{blue} and indoor experiments with a real robot} to demonstrate safe robot motion that can influence the surrounding humans. We also validate the trajectory forecasting performance of \ac{ORCA} on a human trajectory dataset.
Code: \href{https://github.com/sepsamavi/safe-interactive-crowdnav.git}{\footnotesize \texttt{github.com/sepsamavi/safe-interactive-crowdnav.git}}.
{ }\\

\end{abstract}


\section{Introduction} \label{sec:intro}
\begin{figure}[t]
    \footnotesize
    \centering
    \includegraphics[width=\linewidth]{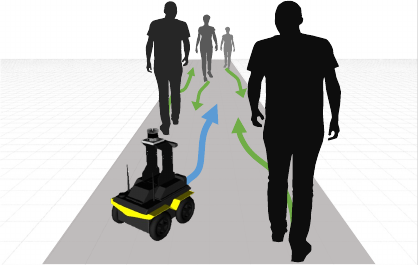}
    \caption{Our MPC algorithm, SICNav, jointly optimizes robot trajectories (blue) and associated predicted human trajectories (green) for multiple humans while satisfying kino-dynamic and collision-avoidance safety constraints. {\color{blue} Video of our work deployed on the pictured robot: \href{https://tiny.cc/sicnav_overview}{\texttt{tiny.cc/sicnav\_overview}}}.}
    \label{fig:problem_overview}
\end{figure}

For humans, navigating within crowds is considered a trivial task. However, uncertainty in the intent and future motions of pedestrians poses considerable challenges to safe (i.e. collision-free) robotic crowd navigation (e.g. in an environment similar to Fig.~\ref{fig:problem_overview}).

In a {\color{blue} classical} approach, the robot decouples motion prediction from planning. It first uses a prediction model (e.g. \cite{saltzmann2021trajpp,mangalam2021ynet,yue2022nspsfm}) in open-loop to forecast the motion of humans, then uses these forecasts to generate robot actions that avoid future collisions  (e.g \cite{Fox1997dwa,Khatib1997eb,DuToit2012,Rosmann2017teb}).
While these approaches are safe by design, e.g. by providing rigorous constraint-satisfaction guarantees \cite{mayne2000constMPC,DuToit2012}, they neglect the fact that the humans and the robot form a closed-loop system; the agents influence each other and need to \emph{interactively} coordinate movement. Recent work has even shown that, in a decoupled framework, using a prediction model with high accuracy to replace one with lower accuracy, does not lead to better robot navigation performance \cite{poddar2023fromcrowdmotiontorobot}. Furthermore, these approaches have been shown to exhibit the \ac{FRP}~\cite{Trautman2010,DuToit2012}, where predictive uncertainty grows to the point that the robot cannot compute a safe path with sufficient certainty.

Recognizing the limitations of such decoupled strategies has led to the development of more sophisticated closed-loop prediction and planning methods to jointly generate predictions and robot motion. Some methods explicitly model cooperation (e.g. \cite{Trautman2015,sun_move_2021}) or competition (e.g. \cite{Sadigh2016}) among agents, while others use \ac{RL} techniques to learn policies that implicitly reason about interactions (e.g. \cite{Chen2019c,chen_relational_2020,Everett2021}). Despite these advancements, existing interactive solutions are not designed to satisfy explicit safety constraints, and either couple the often opposing objectives of safety and performance (e.g. in collision-avoiding \ac{RL} \cite{Everett2021}), or need to rely on post-processing to ensure safety (e.g. via safety filters \cite{cheng2019rlcbf}).

Towards addressing this gap, we present \ac{SICNav}, a nonlinear \ac{MPC} approach for jointly generating closed-loop predictions and planning robot actions (illustrated in green and blue in Fig.~\ref{fig:problem_overview}, respectively). \ac{SICNav} explicitly models interactions between the robot and human agents, but unlike the aforementioned generative methods, allows for safety to be formulated as explicit constraints on the state space.
We model each human in the environment as an agent following the \ac{ORCA} algorithm \cite{vandenBerg2011orca}.
Although originally intended for decentralized control of a swarm, \ac{ORCA} is also commonly used to simulate human crowds for training \ac{RL} agents \cite{Chen2019c, Liu2021socialnce} and testing a variety of other methods \cite{Katyal2020, sun_move_2021}.
Furthermore, \ac{ORCA} has even been used to efficiently generate open-loop predictions for humans in multi-agent environments \cite{Chen2022orcampc}, demonstrating comparable predictive accuracy as state-of-the-art methods (e.g. \cite{Alahi2016a,Ivanovic2018multimultipred}).

Our key insight is to directly incorporate \ac{ORCA} into a robot planner.
We formulate \ac{SICNav} as a \textit{bilevel} optimization problem. The \textit{upper-level} contains the robot's objective, safety constraints, and dynamics constraints. We add a set of human prediction constraints (one for each human), each of which is the solution to a \textit{lower-level} \ac{ORCA} optimization problem. This way, the predictions used for planning are closed-loop: our algorithm jointly generates robot actions and associated predicted human motion by assuming that the humans solve \ac{ORCA} in reaction to robot motion \cite{Dragan2017planning}.

The main goal of this paper is to present the theoretical foundations of our formulation and validate that they are sound, in preparation for the key next step of deploying our approach with real humans {\color{blue} in the wild}. Our contributions are {\color{blue} fourfold}.
We first formulate the bilevel SICNav optimization problem. To solve, we derive a KKT-reformulation of the bilevel problem by replacing the lower-level \ac{ORCA} problems with their \ac{KKT} conditions, which we obtain through Lagrangian duality, and present a theoretical analysis into the equivalence of solving the reformulation versus the original problem.
Second, to justify the use of \ac{ORCA} as a model of pedestrians, we validate its trajectory forecasting performance on real human data in the commonly-used ETH/UCY benchmark \cite{lerner2007ucydata,pellegrini2009ethdata}.
{\color{blue} Third}, we conduct an experimental analysis of SICNav in {\color{blue} two simulation environments} and observe interactive behavior between the robot and simulated humans, such as waiting for the other agent to pass and persuading another agent to move out of the way.
Finally, {\color{blue}  we evaluate SICNav on a real robot in a lab environment, demonstrating improved navigation performance compared to a baseline method and making qualitative observations a robot using SICNav to navigate.}

\section{Related Work} \label{sec:related_work}
\emph{Open-loop prediction: }
This paradigm \textit{decouples} predictions about the future from motion planning.
The robot's planner takes as input the predicted future trajectories of pedestrians and reactively generates a plan to go around the predicted human motion (e.g. \cite{Kummerle2013, Kretzschmar2016, DuToit2012, Katyal2020, Thomas2021}).
The consequence of this decoupling is that the impact of the robot's motion on the predicted trajectory of the humans is assumed to be static with respect to the robot's plan.
Inaccurate predictions may lead to unsafe behavior, and in dense multi-human scenarios, high predictive uncertainty can result in the~\ac{FRP}~\cite{Trautman2010}, where the robot freezes because the planner cannot find a path that is sufficiently certain to be collision-free. Furthermore, following this paradigm neglects the fact that the robot may be able to influence pedestrians to be cooperative and allow the robot to avoid freezing \cite{Dragan2017planning}.
In contrast to these methods, we move away from the \textit{predict-then-plan} paradigm by modelling the interactions between the robot and pedestrians.
As our optimization algorithm is searching for an optimal solution for the robot's trajectory, it concurrently optimizes predictions of the optimal human actions (with respect to \ac{ORCA}) given the current iteration of the robot actions being optimized. Thus, our predictions adapts to the plan being evaluated.

\emph{Cooperative interactions: }Another family of approaches assumes that all agents are cooperative in avoiding collisions. For example, the robot can model all agent trajectories (including its own) as \acp{GP} that represent a distribution of trajectories for each agent, then define an `interactive' cost to minimize the likelihood that trajectory distributions will overlap (i.e. collide) in belief space \cite{Trautman2015, sun_move_2021}. However, the assumption that all agents are exclusively cooperative may not always hold true. In many crowded scenarios, such as a densely packed urban sidewalk, pedestrians may compete for space and try to influence each other's behavior to arrive at their destinations.

\emph{Competitive interactions: }A number of recent methods aimed at self-driving capture a similar competition in highway merging by casting \ac{AV} navigation among human-operated cars as a dynamic zero-sum game, then use a variety of approximations to optimize the robot's reward without needing to solve for a Nash equilibrium \cite{Sadigh2016} {\color{blue} \cite{Ren2019, Fisac2019, kedia2023gametheoretic}}.
The humans are modelled as \textit{myopic} non-cooperative agents, i.e. the human and the robot have different reward functions, but the human is modelled to be following the lead of the robot, i.e. humans calculate the best response to the robot's actions \cite{Sadigh2016, Schwarting2019}. This formulation, usually referred to as a Stackelberg game, has demonstrated interesting interactive behavior such as a robot moving in reverse to signal a human to move first at an intersection. However, these works usually focus on one-on-one driving scenarios, which may limit their applicability to navigating pedestrian crowds. First, crowds are often comprised of a variable number multiple agents all interacting with each other. Second, the driving environment is much more structured than pedestrian environments that usually lack explicit navigation rules. Similar to the two-agent game theoretic approaches, our approach models humans as \textit{myopic} optimal agents following an \ac{ORCA} strategy, but our planner uses bilevel optimization to output locally optimal plans in less structured crowd navigation settings while reasoning about multiple humans.

\section{Problem Formulation and Background}
\begin{figure*}[th]
	\centering
	\footnotesize
	\begin{subfigure}[]{0.33\textwidth}
		\centering
	    \footnotesize
		\includegraphics[width=\textwidth]{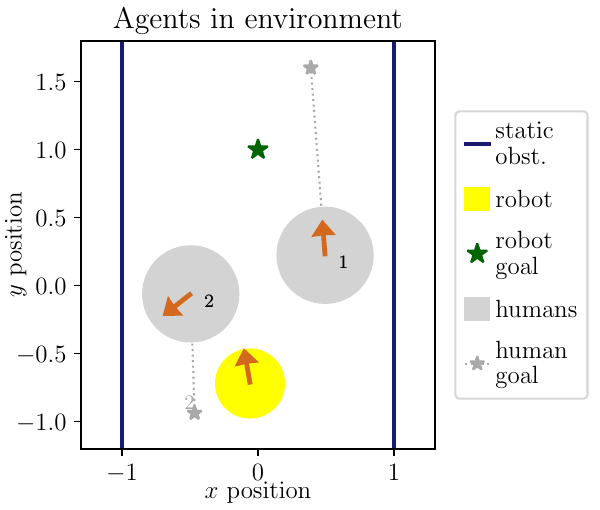}
		\caption{Crowd navigation environment}
	    \label{fig:env_setup_posns}
	\end{subfigure}
	\hfill
	\begin{subfigure}[]{0.66\textwidth}
		\centering
	    \footnotesize
		\includegraphics[width=\textwidth]{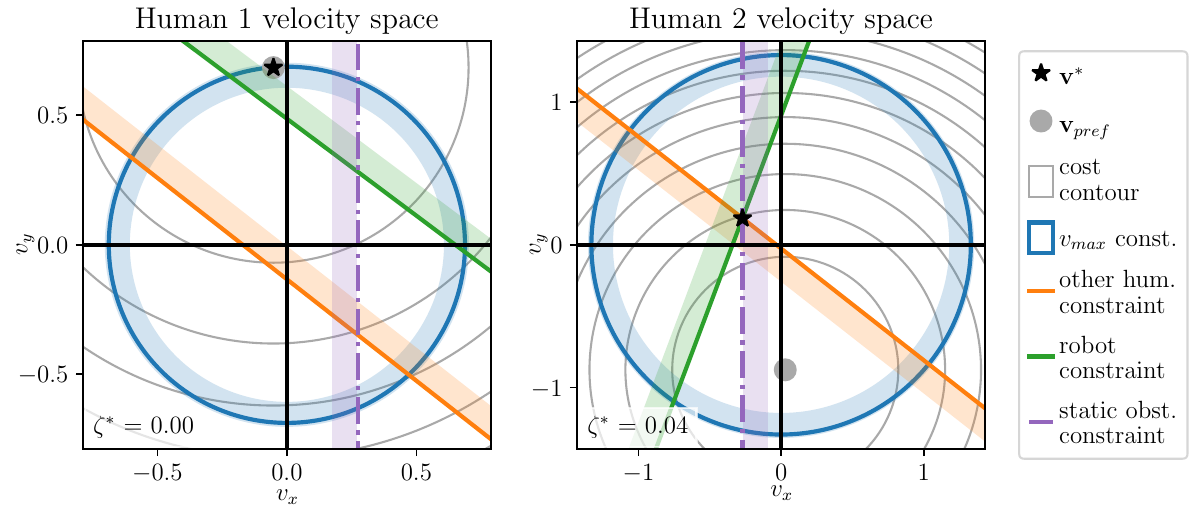}
		\caption{\ac{ORCA} optimization}
	    \label{fig:env_setup_vels}
	\end{subfigure}
    \caption{(a) Exemplary crowd navigation environment state $\stateat{t}$ with one robot (yellow) and its goal position (green star), $\numhumans=2$ simulated humans (gray circles) with respective goals (gray stars), agent velocities (red arrows), and $\numstatobs = 2$ static obstacles (blue lines). (b) Graphical representation of the associated \ac{ORCA} velocity space optimization \eqref{eq:relaxed_orca_argmin} solved by each simulated human (Human 1 left and Human 2 right). The human's preferred velocity (gray circle) points toward their goal, and induces the minimum of the cost (gray contours). The constraints are illustrated as lines (for non-collision constraints) or a circle (for the maximum velocity constraint) with the feasible sides shaded. The optimum velocity, $\vel^*$, i.e. the closest velocity to $\vel_{\pref}$ that is feasible with respect to every constraint, is illustrated as a black star, with the associated optimal value of the slack variable, $\slackvar^*$, written on the graph.}
	\label{fig:env_setup}
\end{figure*}
The environment consists of a robot, human agents, indexed $\idA \in \forallhumanslongset$, and static obstacles in the form of line segments and indexed, $\idStat \in \forallstatobslongset$. Fig.~\ref{fig:env_setup_posns} illustrates an initial configuration consisting of one robot (in yellow) and $\numhumans=2$ simulated human agents (in gray).
The goal of the robot is to move to a goal location (green star in Fig.~\ref{fig:env_setup_posns}), $\goal$, starting from an initial configuration, $\stateinit$, without any collisions with static obstacles or simulated human agents.

\subsection{Robot and Human System Dynamics} \label{sec:sysstatedyn}
The system state is continuous and concatenates the states of all agents,
one robot and $\numhumans$ humans. At time step $t$, the state is,
	$
    \stateat{t} = (\rob{\stateat{t}}, \id{1}{\stateat{t}}, \dots, \id{\numhumans}{\stateat{t}})  \in \sspace,
	$
where the state of the robot, {\color{blue} $\rob{\stateat{t}} \in \Reals^{4}$, consists of its 2D position, heading, linear velocity at $t-1$}, and the state of each human agent, $\id{\idA}{\stateat{t}} \in \Reals^{4}$, consists of its 2D position, and 2D velocity. {\color{blue} We also assume to know the 2D goal position of each human. The goal of other agents is privileged information that is unavailable in the real-world, in Sec.~\ref{sec:sim_results} we discuss how this assumption can be relaxed when deploying the algorithm.}

We separate the dynamics of the system into separate functions for the robot and the humans. The robot dynamics,
$\rob{\stateat{\kpone}} = \robdyn(\rob{\stateat{t}}, \actionat{t}),$
are modeled as a kinematic unicycle model, where the control input for the system, $\at{\action}{t} \in \cspace$, is a vector of the linear and angular velocity of the robot for time step $t$.
The dynamics of each human $\forallhumans$,
$\id{j}{\stateat{\kpone}} = \humdyn(\id{j}{\stateat{t}}, \id{\idA}{\at{\actionhum}{t}}),$
are modeled as kinematic integrators, where $\id{\idA}{\at{\actionhum}{t}}$ denote the human actions. Note that these actions are not an input to the system, but rather predictions of the human actions given the state of the system, $\stateat{t}$.

\subsection{ORCA Model of Human Motion}
We obtain these predictions of human actions at each step, $\id{\idA}{\at{\actionhum}{t}}$, by solving the \ac{ORCA} optimization problem \cite{vandenBerg2011orca}. Before defining the complete optimization problem, we introduce a basic version. For one agent $\idA$ at one time step $t$, we have the following set of optimal (i.e. minimum-cost feasible) velocities,
\begin{equation} \label{eq:orca_lp1_argmin}
    \begin{alignedat}{2}
      \centremathcell{{\orcasolnset}(\at{\state}{t}) :=}	& \centremathcell{\argmin_{\vel}}	& \centremathcell{\twonorm{\vel - {\vel_{\pref}}(\stateat{t})}^2}\\
      \centremathcell{}												& \centremathcell{\subjectto\text{ }}					& \centremathcell{\twonorm{\vel}^2 \leq {\lvel_{\max}}^2} \\
      \centremathcell{}												& \centremathcell{}										& \centremathcell{\id{\idA|\idB}{\orcalinedir}(\stateat{t})^\top \vel \geq \id{\idA|\idB}{\orcaconstgeq}(\stateat{t})}
    \end{alignedat}
\end{equation}
where the optimization variable $\vel \in \Reals^2$ is the velocity of agent $j$ and the preferred velocity ${\vel_{\pref}} : \sspace \to \Reals^2$ is a vector toward the agent's goal position.
The maximum velocity of the agent is constrained by ${\lvel_{\max}}$, and the vectors $\id{\idA|\idB}{\orcalinedir} : \sspace \to \Reals^2$, which are piece-wise linear functions of $\stateat{t}$, and scalar $\id{\idA|\idB}{\orcaconstgeq} : \sspace \to \Reals^2$ which are nonlinear functions of $\stateat{t}$, define linear collision avoidance constraints between agents $j$ and agent $l, \forallorcas$. The cost and constraint functions in \eqref{eq:orca_lp1_argmin} depend on the positions and velocities of all agents in the system at each time step, therefore, the state of the system, $\stateat{t}$, parameterizes this optimization problem. In the remainder of this section, we will reference Fig.~\ref{fig:env_setup}, which illustrates an exemplary environment state (Fig.~\ref{fig:env_setup_posns}) and a graph of the associated \ac{ORCA} problem solved by each human (Fig.~\ref{fig:env_setup_vels}).

\emph{Agent-agent collision avoidance}: The normal vectors, $\id{\idA|\idB}{\orcalinedir}({\at{\state}{t}})$, and constants, $\id{\idA|\idB}{\orcaconstgeq}({\at{\state}{t}})$, are formulated using a \ac{RVO} approach \cite{vandenBerg2011orca}. Assume agent $\idA$ is solving \eqref{eq:orca_lp1_argmin}. For each other agent, $\idB \neq \idA$ (e.g. the robot and human 2 in Fig.~\ref{fig:env_setup}), agent $\idA$ uses the current relative position and velocity between itself and $\idB$ to compute a region of relative velocity space, called a \emph{velocity obstacle} \cite{fiorini1998vo}, which contains all relative velocities that would cause a collision between the agents within some designated time horizon $\colTau$. Assuming that both agents follow the same velocity-obstacle protocol, the algorithm finds uses this region to find a half-plane in the velocity space of $\idA$ that forbid the agent from choosing a velocity for the next time step that would cause a collision with the other agent within $\colTau$.
In the case of human $\idA=1$ in Fig~\ref{fig:env_setup_vels}, the green half plane with the feasible side shaded is induced by the position and velocity of the robot$=\idB$. As illustrated in Fig.~\ref{fig:env_setup_posns}, the robot is below and to the left of human 1 and moving upwards ($y\uparrow$), causing the velocity space of human 1 to be forbidden from velocities that move it to the left ($x\downarrow$) or downwards ($y\downarrow$) as that may cause a collision with the robot within $\colTau$. Likewise, the position and velocity of human $\idB=2$ induces the orange constraint for human 1.

\emph{Static obstacle collision avoidance}:
The algorithm uses a simplified \ac{RVO} derivation to obtain half-plane constraints that avoid static obstacles defined as line segments, e.g. the static obstacles on the right and left in Fig.~\ref{fig:env_setup_posns}. These constraints are similarly represented with a normal vector, $\id{\idA|\idStat}{\orcalinedir}({\at{\state}{t}})$, and a scalar, $\id{\idA|\idStat}{\orcaconstgeq}({\at{\state}{t}})$, for all static obstacle indices $\forallstatorcas$. In the velocity space of human 1 in Fig.~\ref{fig:env_setup_vels}, the static obstacle on the right in Fig.~\ref{fig:env_setup_posns} induces the dashed purple constraint to forbid human 1 from selecting a velocity towards the right ($x \uparrow$) that would cause a collision with the obstacle.

\emph{Feasible velocities}:
For human 1 in Fig.~\ref{fig:env_setup}, the feasible set of velocities are formed by the aforementioned non-collision constraints for the static obstacle and the robot, together with human 1's maximum permissible velocity (blue circle). In the graph for human 1 in Fig.~\ref{fig:env_setup_vels}, we can see that the feasible set of velocities as the intersection of the shaded sides of these three constraints (green, blue and purple). In the case of human 1, the preferred velocity lies in this feasible set and is the optimum.

\emph{Empty feasible set and ORCA relaxation}:
As discussed in \cite{vandenBerg2011orca}, in dense scenarios i.e. where agents are in close vicinity to each other, the problem \eqref{eq:orca_lp1_argmin} may become infeasible, i.e. the feasible set does not contain any points.
In this case, the \ac{ORCA} algorithm \cite{vandenBerg2011orca} switches to a secondary optimization problem that finds the safest-possible feasible velocity by minimizing the maximum distance of the velocity variable $\vel$ to the half-plane $\id{\idA|\idB}{\orcalinedir}(\stateat{t})^\top \vel \geq \id{\idA|\idB}{\orcaconstgeq}(\stateat{t})$ (See Sec. 5.2 of \cite{vandenBerg2011orca}). The secondary problem is a relaxation equivalent to moving every pairwise ORCA half-plane outward (i.e. opposite direction to the feasible side) equally, until at least one velocity becomes feasible, then choosing the lowest magnitude velocity. We formulate a similar relaxation by adding a slack variable to the linear collision avoidance constraints in \eqref{eq:orca_lp1_argmin}. We only add slack variables for the agent-agent collision constraints as only those constraints may result in infeasibility. Putting the relaxation and static obstacle constraints together, we obtain the following optimization problem, which we will use as our prediction model for humans in \ac{SICNav}.

\begin{definition}[Relaxed ORCA]
	Let ${\slackvar}$ be a slack variable, and let $\penal \in \Reals_{++}$ be some sufficiently large constant. The \emph{set of velocities solving the relaxed \ac{ORCA} problem for agent $\idA$} is,
	\begin{equation} \label{eq:relaxed_orca_argmin}
		\begin{alignedat}{2}
		\centremathcell{{\orcarlxsolnset}(\at{\state}{t}) :=}	& \centremathcell{\argmin_{\varsoneorca}}	& \centremathcell{\twonorm{\vel- {\vel_{\pref}}(\stateat{t})}^2 + \penal {\slackvar}^2}\\
		\centremathcell{}													& \centremathcell{\subjectto\text{ }}												& \centremathcell{\twonorm{\vel}^2 \leq {\lvel_{\max}}^2} \\
		\centremathcell{}													& \centremathcell{}																	& \centremathcell{\id{\idA|\idB}{\orcalinedir}(\stateat{t})^\top \vel \geq \id{\idA|\idB}{\orcaconstgeq}(\stateat{t}) - {\slackvar}}\\
		\centremathcell{}													& \centremathcell{}																	& \centremathcell{\id{\idA|\idStat}{\orcalinedir}(\stateat{t})^\top \vel \geq \id{\idA|\idStat}{\orcaconstgeq}(\stateat{t})}\\
		\centremathcell{}													& \centremathcell{}																	& \centremathcell{{\slackvar} \geq 0}
		\end{alignedat}
	\end{equation}
\end{definition}
The relaxed problem is a convex three-dimensional \ac{QCQP}. We use an inexact penalty function to relax the agent-agent collision constraints on the cost function. The upper bound of the error tends to $0$ as $\penal \to \infty$.
Unlike the original paper \cite{vandenBerg2011orca} which picks the smallest velocity after relaxing the constraints, our approach picks the one closest to the preferred velocity.

The slack variable can be visualized as a third dimension protruding out of the page in Fig.~\ref{fig:env_setup_vels}. In the case of human 2 there is \textbf{no} feasible velocity for $\slackvar=0$ (not visualized), thus the agent-agent collision constraints are relaxed to $\slackvar^*=0.04$ (visualized), and we observe the single feasible point (black star) that becomes feasible is selected as the optimum. Since the value of $\penal$ is large, increasing the value of $\slackvar$ is severely penalized, meaning that the constraints are relaxed minimally (i.e. least unsafe) in order to obtain at least one feasible velocity. To discern the consequences of the relaxation, recall that \ac{ORCA} without the relaxation \eqref{eq:orca_lp1_argmin} is guaranteed to be collision-free if the agents follow their optimal velocities for some designated time horizon, $\colTau$. An optimal velocity with $\slackvar>0$ means that a collision may occur within time $\colTau$. However, this time horizon is set to be much larger than one typical time step, e.g. $\colTau=2.0s$ in Fig.~\ref{fig:env_setup}, therefore it is very unlikely that a collision ever occurs in practice.
\section{SICNav MPC Problem} \label{sec:our_soln}
\begin{figure}[t]
	\centering
	\footnotesize
	\includegraphics[width=\linewidth]{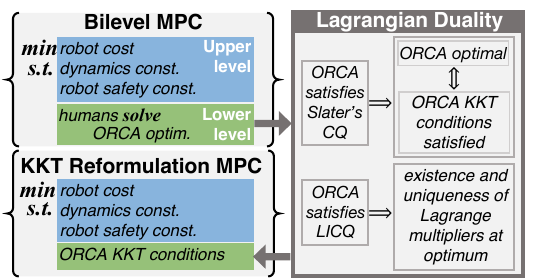}
    \caption{Overview of our method. In section~\ref{sec:our_soln}, we formulate a Bilevel MPC problem \eqref{eq:combmpc0_prob} that solves for robot actions in the upper level (blue) and predicted human actions via a lower level \ac{ORCA} problems (green). In section~\ref{sec:orca_properties}, we use Lagrangian duality analysis to derive the \ac{KKT} optimality criteria for the lower level problems, and in section~\ref{sec:kkt_reform}, we reformulate the bilevel problem by replacing the lower level problems with their \ac{KKT} optimality criteria to get a single level KKT Reformulation \ac{MPC} problem that we can solve. In section~\ref{sec:kkt_reform_properties}, we analyze the equivalence between optima of the reformulation and optima of the original bilevel problem.}
	\label{fig:overview}
\end{figure}
Fig.~\ref{fig:overview} provides a graphical overview of our approach described in this section and the next.
We formulate an \ac{MPC} point stabilization controller as a bilevel optimization with the \ac{ORCA} problem \eqref{eq:relaxed_orca_argmin} as the lower level.

\emph{Robot cost:} We define the components of the stage cost to penalize deviation from the robot's goal {\color{blue} state, $\error_t := \stateat{t} - \state_g$}, and excessive control effort,
{\color{blue}%
\begin{equation*} \label{eq:mpc_cost}
	\stagecostsymb(\stateat{t}, {\actionat{t}}) = \trans{\error_t} \Q \error_t + \trans{\actionat{t}} \R \actionat{t},
\end{equation*}
with positive semidefinite matrices $\Q$ and $\R$. We also define a terminal penalty as,
\begin{equation*} \label{eq:mpc_termcost}
	\termpenal{\stateat{\horiz}} = \beta \trans{\error_{\horiz}} \Q \error_{\horiz},%
\end{equation*}}
where $\beta \geq 1$ is selected to be sufficiently large to ensure stability of the controller (See Theorem 2.41 in \cite{rawlings2019mpcbook}). {\color{blue} In practice, we select the goal state, $\state_g$ to include the robot's final goal pose and target linear velocity, and we select $\Q$ to be diagonal and penalize only the robot's state, not the predicted state of the humans. We select $\R$ to be diagonal and penalize the control effort of the robot.}

\emph{Robot safety constraints}: To avoid collisions between the robot and any human agents we add constraints of the form,
\begin{equation*} \label{eq:rob_coll_const}
	\trans{\at{\state}{t}} \P_\idA \at{\state}{t} \geq \dist_\idA^2, \forallhumans,
\end{equation*}
where, $\P_\idA \in \Reals^{n \times n}$ is a matrix that extracts the positions of the robot and each human agent $\idA$ from the state and $\dist_\idA$ is the minimum permissible distance between the two agents.
For each static obstacle, we implement a piece-wise function that calculates the closest point on the line segment to the position of the robot and add a quadratic constraint similar to the robot-agent collision constraints.
We also bound the input of the system in order to meet the kino-dynamic limits of a real-world robot,
\begin{equation*}
    \begin{aligned}
	   \centremathcell{\id{\idRob}{\action_{\min}} \leq \at{\action}{t} \leq \id{\idRob}{\action_{\max}},} \\
        \centremathcell{\Delta\id{\idRob}{\action_{\min}} \leq \at{\action}{t} - \at{\action}{\kmone} \leq \Delta\id{\idRob}{\action_{\max}},}
    \end{aligned}
\end{equation*}

With these definitions, the local planning \ac{MPC} problem for the robot can be formulated as an optimization problem,
\begin{subequations} \label{eq:combmpc0_prob}
	\begin{alignat}{2}
		\centremathcell{\minimize_{\substack{\stateat{0:\horiz},\actionat{0:\horiz-1},\\\id{0:\numhumans}{\at{\actionhum}{0:\horiz-1}}}}} 	& \centremathcell{\sum_{t=0}^{\horiz-1} \stagecost{\stateat{t},\actionat{t}} + \termpenal{\stateat{\horiz}}} 		& \centremathcell{\quad\quad\quad\quad\quad} \label{eq:combmpc0_min}			\\
		\centremathcell{\subjectto} 																			  							& \centremathcell{\stateat{0}=\stateinit} 																	 		& \centremathcell{\quad\quad\quad\quad\quad} \label{eq:combmpc0_initcondconst}	\\
		\centremathcell{}																										 			& \centremathcell{\rob{\stateat{\kpone}}= \robdyn(\rob{\stateat{t}},\controlat{t})} 							     		& \centremathcell{\quad\quad\quad\quad\quad} \label{eq:combmpc0_rob_dynconst}	\\ 	
		\centremathcell{}																										 			& \centremathcell{\id{\idRob}{\action_{min}} \leq {\at{\action}{t}} \leq \id{\idRob}{\action_{\max}}} 	 		& \centremathcell{\quad\quad\quad\quad\quad} \label{eq:combmpc0_actionconst0}	\\ 	
		\centremathcell{}																										 			& \centremathcell{\Delta\id{\idRob}{\action_{min}} \leq {\at{\action}{t} - \at{\action}{\kmone}} \leq \Delta\id{\idRob}{\action_{\max}}} 	 		& \centremathcell{\quad\quad\quad\quad\quad} \label{eq:combmpc0_actionconst}	\\ 	
		\centremathcell{}																										 			& \centremathcell{\trans{\at{\state}{t}} \P_\idB \at{\state}{t} \geq {\dist_{\idB}}^2}   	& \centremathcell{\quad\quad\quad\quad\quad} \label{eq:combmpc0_coll_const}	\\ 	
	  	\centremathcell{}																										 			& \centremathcell{\id{\idA}{\at{\actionhum}{t}} \in \id{\idA}{\orcarlxsolnset}(\at{\state}{t})}				 		& \centremathcell{\quad\quad\quad\quad\quad} \label{eq:combmpc0_llorca}		\\ 	
	  	\centremathcell{}																										 			& \centremathcell{\id{\idA}{\stateat{\kpone}} = \humdyn(\id{\idA}{\stateat{t}}, \id{\idA}{\at{\actionhum}{t}})}	 		& \centremathcell{\quad\quad\quad\quad\quad} \label{eq:combmpc0_hum_dynconst}		
	\end{alignat}
\end{subequations}
where all the constraints \eqref{eq:combmpc0_rob_dynconst}-\eqref{eq:combmpc0_hum_dynconst} are defined for each time step, $\forallactidcs$, and the constraints \eqref{eq:combmpc0_llorca}-\eqref{eq:combmpc0_hum_dynconst} are defined for each human, $\forallhumans$. The robot collision constraint \eqref{eq:combmpc0_coll_const} is defined for each human agent and each static obstacle in the environment, $\forall \idB \in \set{1,\dots,\numhumans,\dots,\numhumans+\numstatobs}$.
To solve this bilevel problem, we will reformulate to a single level by replacing the lower level with its \ac{KKT} optimality conditions.
As such, we begin by characterizing the optimality conditions for the \ac{ORCA} problem \eqref{eq:relaxed_orca_argmin}.

\subsection{Lagrangian Duality Analysis of ORCA Problem} \label{sec:orca_properties}
To analyze the optimal solutions of the \ac{ORCA} problem \eqref{eq:relaxed_orca_argmin}, we use Lagrange duality theory. First, we define the Lagrangian of the \ac{ORCA} problem \eqref{eq:relaxed_orca_argmin}. Recall that each agent $\idA$ has its own instance of the problem, however in the remainder of this section we drop the index $\idA$ for legibility. %
\begin{definition}[Lagrangian Dual]
	\label{def:lag_dual}
	For the \ac{ORCA} problem associated to agent $\idA$ \eqref{eq:relaxed_orca_argmin}, let the optimization variable be $\varoneorca = \varsoneorcastacked \in \Reals^3$, let the cost function be, %
	${\varphi}(\varoneorca; \at{\state}{t})$, %
	let the constraints be summarized by ${\mathbf{c}}(\varoneorca; \at{\state}{t}) \leq \mathbf{0}$, and let $\lagmuls \in \Reals^{\numhumans+\numstatobs+2}$. %
	The \emph{Lagrangian} of problem \eqref{eq:relaxed_orca_argmin} is,
	\begin{equation} \label{eq:orca_lagrangian}
			\lagfun(\varoneorca, \lagmuls;\at{\state}{t}) =
			\varphi(\varoneorca;\at{\state}{t}) + \lagmuls^\top {\mathbf{c}}(\varoneorca ; \at{\state}{t})
	\end{equation}
	where $\lagmuls \geq \mathbf{0}$ are the \emph{Lagrange multipliers}.
	The \emph{dual optimization} of the problem \eqref{eq:relaxed_orca_argmin} is,
	\begin{equation} \label{eq:orca_dual_prb}
	\max_{\lagmuls}\inf_{\varoneorca} \lagfun(\varoneorca, \lagmuls;\at{\state}{t}).
	\end{equation}
\end{definition}
For there to exist some $\lagmuls^*$ such that the dual optimization is maximized, the problem \eqref{eq:relaxed_orca_argmin} needs to satisfy the following constraint qualification \cite{boyd2004convex}.
\begin{definition} [\acs{SCQ}] \label{def:slaters}
	\emph{\ac{SCQ}} is satisfied for problem \eqref{eq:relaxed_orca_argmin} if there exists a point $\varoneorca$ in the interior of the feasible set such that ${\mathbf{c}}(\varoneorca ; \at{\state}{t}) < \mathbf{0}$.
\end{definition}

Problem \eqref{eq:relaxed_orca_argmin} satisfies \acs{SCQ} by construction. Starting at any point, $\varoneorca$, the value of the slack variable dimension, ${\slackvar}$, can be increased such that the pairwise collision constraints are relaxed sufficiently for the feasible set to have an interior.
\begin{definition}[KKT Conditions]
	The \emph{\acl{KKT} (\acs{KKT})} conditions of \eqref{eq:relaxed_orca_argmin} are defined as \cite{boyd2004convex},
	\begin{subequations} \label{eq:orca_kkt_conds}
		\begin{alignat}{2}
			& \text{\emph{Stationarity:}      }&{ }& \centremathcell{\nabla_{\varoneorca} \lagfun(\varoneorca, \lagmuls ; \at{\state}{t}) = \mathbf{0}} \label{eq:orca_kkt_stat} \\
			& \text{\emph{Complementary Slackness:}   }&{ }& \centremathcell{\lagmuls^\top {\mathbf{c}}(\varoneorca ; \at{\state}{t}) = 0} \label{eq:orca_kkt_compslack} \\
			& \text{\emph{Primal Feasibility:}}&{ }& \centremathcell{{\mathbf{c}}(\varoneorca ; \at{\state}{t}) \leq \mathbf{0}} \label{eq:orca_kkt_primfeas} \\
			& \text{\emph{Dual Feasibility:}  }&{ }& \centremathcell{\lagmuls \geq \mathbf{0}} \label{eq:orca_kkt_dualfeas}
		\end{alignat}
	\end{subequations}
\end{definition}
Since the problem is convex and satisfies \ac{SCQ},
the \ac{KKT} conditions are necessary and sufficient for global optimality of the problem \eqref{eq:relaxed_orca_argmin}\cite{boyd2004convex}.
Furthermore, note that the cost function in problem \eqref{eq:relaxed_orca_argmin} is strongly convex with respect to $\varoneorca$ due to the quadratic form of its cost function. Thus, the set ${\orcarlxsolnset}(\at{\state}{t})$ is always singleton, i.e. the solution to the optimization problem is always unique\cite{boyd2004convex}.
In the next section, we will make use of the \ac{KKT} conditions and the uniqueness of solutions reformulate a bilevel optimization, with problem \eqref{eq:relaxed_orca_argmin} as the lower-level, to a single level problem.

\subsection{KKT Reformulation of bilevel problem} \label{sec:kkt_reform}
In order to solve \eqref{eq:combmpc0_prob}, we replace each instance of the lower level problem with its \ac{KKT} conditions \eqref{eq:orca_kkt_conds}, to obtain the following \ac{MPCC},
\begin{subequations}\label{eq:combmpckkt_prob}
	\begin{alignat}{3}
		\centremathcell{\minimize_{\substack{\stateat{0:\horiz}, \actionat{0:\horiz-1},\at{\id{0:\numhumans}{\actionhum}}{0:\horiz-1},\\
		\id{0:\numhumans}{\at{\slackvar}{0:\horiz-1}},\id{0:\numhumans}{\at{\lagmuls}{0:\horiz-1}}}}} & \centremathcell{  \sum_{t=0}^{\horiz-1} \stagecost{\stateat{t},\actionat{t}} + \termpenal{\stateat{\horiz}}} 																										&\centremathcell{\quad\quad\quad\quad\quad\quad\quad} \label{eq:combmpckktkkt_min} \\
		\centremathcell{\text{subject to\quad}}															  & \centremathcell{  \stateat{0}=\stateinit}  																																										&\centremathcell{\quad\quad\quad\quad\quad\quad\quad} \label{eq:comb_mpc_initcondconst} \\
	  																								   { }& \centremathcell{  \rob{\stateat{\kpone}}=\robdyn(\stateat{t},\controlat{t})} 																																	&\centremathcell{\quad\quad\quad\quad\quad\quad\quad} \label{eq:combmpckkt_rob_dynconst} \\ 
	  																								   { }& \centremathcell{  \id{\idRob}{\action_{min}} \leq  \rob{\at{\action}{t}} \leq \id{\idRob}{\action_{\max}}} 																											&\centremathcell{\quad\quad\quad\quad\quad\quad\quad} \label{eq:combmpckkt_actionconst0} \\ 
																									   { }& \centremathcell{  \Delta\id{\idRob}{\action_{min}} \leq  \rob{\at{\action}{t}}-\rob{\at{\action}{\kmone}} \leq \Delta\id{\idRob}{\action_{\max}}} 																											&\centremathcell{\quad\quad\quad\quad\quad\quad\quad} \label{eq:combmpckkt_actionconst} \\ 
																									   { }& \centremathcell{  \trans{\at{\state}{t}} \P_\idB \at{\state}{t} \geq \dist_\idB^2} 																										&\centremathcell{\quad\quad\quad\quad\quad\quad\quad} \label{eq:combmpckkt_coll_const} \\ 
																									   { }& \centremathcell{  \nabla_{\id{\idA}{\at{\actionhum}{t}},\id{\idA}{\slackvar}} \lagfun(\id{\idA}{\at{\actionhum}{t}},\id{\idA}{\slackvar},\id{\idA}{\at{\lagmuls}{t}} ; \at{\state}{t}) = \mathbf{0}}	&\centremathcell{\quad\quad\quad\quad\quad\quad\quad} \label{eq:combmpckkt_kktstat} \\ 
																									   { }& \centremathcell{  \id{\idA}{\at{\lagmuls}{t}}^\top \id{\idA}{\mathbf{c}}(\id{\idA}{\at{\actionhum}{t}},\id{\idA}{\slackvar} ; \at{\state}{t}) = 0}												&\centremathcell{\quad\quad\quad\quad\quad\quad\quad} \label{eq:combmpckkt_kktcomp} \\ 
																									   { }& \centremathcell{  \id{\idA}{\mathbf{c}}(\id{\idA}{\at{\actionhum}{t}},\id{\idA}{\slackvar} ; \at{\state}{t}) \leq \mathbf{0}} 																&\centremathcell{\quad\quad\quad\quad\quad\quad\quad} \label{eq:combmpckkt_kktfeas} \\ 
																									   { }& \centremathcell{  \id{\idA}{\at{\lagmuls}{t}} \geq \mathbf{0}} 																																					&\centremathcell{\quad\quad\quad\quad\quad\quad\quad} \label{eq:combmpckkt_kktdualfeas} \\ 
																									   { }& \centremathcell{  \id{\idA}{\stateat{\kpone}} = \humdyn(\stateat{t}, \id{\idA}{\at{\actionhum}{t}})} 																										&\centremathcell{\quad\quad\quad\quad\quad\quad\quad} \label{eq:comb_mpc_hum_dynconst} 
	\end{alignat}
\end{subequations}
where the constraints in \eqref{eq:combmpckkt_prob} defined for each human, $\forall \idA, \forall l, \forall t$ analogously to \eqref{eq:combmpc0_prob}.Each instance of constraint \eqref{eq:combmpc0_llorca} has been replaced by its \ac{KKT} conditions to get \eqref{eq:combmpckkt_kktstat}-\eqref{eq:combmpckkt_kktdualfeas}. Note that the Lagrange multipliers of each instance of \eqref{eq:combmpc0_llorca} have become optimization variables in \eqref{eq:combmpckkt_prob}.

\section{Analysis of KKT-Reformulation} \label{sec:kkt_reform_properties}
We analyze the equivalence of solutions of the KKT reformulation \eqref{eq:combmpckkt_prob} and solutions of the bilevel problem \eqref{eq:combmpc0_prob}. The authors in \cite{dempe2012bilevelmpcc} present sufficient conditions for equivalence.
For globally optimal solution of the reformulated problem \eqref{eq:combmpckkt_prob} to be equivalent to globally optimal solutions of the original \eqref{eq:combmpc0_prob}, it is required that the lower level \eqref{eq:relaxed_orca_argmin} satisfy \ac{SCQ} (Definition~\ref{def:slaters}) for any parameterization $\stateat{t}$ (see Theorem 2.3 in \cite{dempe2012bilevelmpcc}). As discussed in \ref{sec:orca_properties}, the relaxed \ac{ORCA} optimization \eqref{eq:relaxed_orca_argmin} satisfies \ac{SCQ} by construction. Thus, the global optimum of problem \eqref{eq:combmpckkt_prob} is equivalent to the global optimum of \eqref{eq:combmpc0_prob}.

However, problems \eqref{eq:combmpckkt_prob} and the bilevel problem \eqref{eq:combmpc0_prob} are non-convex, therefore we also need to analyze the equivalence of locally optimal solutions. For locally optimal solutions, equivalence cannot be established without imposing further conditions on the lower level problem.
In order for the Lagrange multipliers that optimize the dual problem to be unique, the problem \eqref{eq:relaxed_orca_argmin} needs to satisfy the following additional constraint qualification \cite{wachsmuth2013uniquelaglicq}.
\begin{definition}[LICQ]
	For agent $\idA$ at some environment state, $\at{\bar{\state}}{t}$, which parameterizes the relaxed \ac{ORCA} problem \eqref{eq:relaxed_orca_argmin}, let $\varoneorcalocopt$ be a locally optimal solution for the problem, and let $\ithconstorcarel{i}\varandparamoneorca$ denote the $i^{th}$ row of the constraint function, $\id{\idA}{\mathbf{c}}(\cdot;\cdot)$. Problem \eqref{eq:relaxed_orca_argmin} is said to satisfy the \emph{\ac{LICQ}} if any subset of the indices of active constraints, $I_{a} \subseteq \set{i \st \ithconstorcarelatbar{i} = 0}$, the gradients of the subset of active constraints, $\set{\gradorcavars \ithconstorcarel{i} \varandparamoneorcalocopt \st i \in I_{a}}$, where $\varoneorca = \varsoneorcastacked$, are linearly independent at $\bilevellocoptoneorca$.
\end{definition}

Let the set of solutions to the dual of the lower-level problem \eqref{eq:orca_dual_prb} be defined as,
\begin{equation}
    \Lambda\varandparamoneorca:=\argmax_{{\lagmuls}}\inf_{\varoneorca} \lagfun(\varoneorca, \id{\idA}{\lagmuls};\at{\state}{t}).
\end{equation}
If a candidate solution to \eqref{eq:combmpckkt_prob}, $\varandparamoneorca$, results in a parameterization of the lower-level problem \eqref{eq:relaxed_orca_argmin} that does not satisfy \ac{LICQ}, then the set $\Lambda\varandparamoneorca$ is not singleton, and may even be unbounded \cite{dempe2012bilevelmpcc}.
For our lower level problem \eqref{eq:relaxed_orca_argmin}, we may encounter such a scenario if some $\stateat{t}$ results in two of the linear collision avoidance constraints in \eqref{eq:relaxed_orca_argmin} becoming collinear, and the solution for the lower level problem \eqref{eq:relaxed_orca_argmin} is such that these two constraints are active.
In the worst case, we will be unable to converge to a solution as there will be infinite potential values for the optimal dual variables, $\lagmuls$. In the case that we can converge to a solution, then $(\stateat{t},\varoneorca)$ may be a local optimum of the reformulation \eqref{eq:combmpckkt_prob} but not a local optimum of the bilevel problem \eqref{eq:combmpc0_prob} \cite{dempe2012bilevelmpcc}. Nonetheless, since the \ac{KKT} conditions are necessary and sufficient for the lower-level problem, we know that any local optimum of $(\stateat{t},\varoneorca)$ will be feasible with respect to the bilevel problem.
In the following section we will discuss the closed loop behavior of the planner and how we use a feasible warm start to mitigate the issue of local optima.

\subsection{Feasible Warm Start}
\begin{figure}[!t]
	\centering
    \footnotesize
    \includegraphics[width=\linewidth]{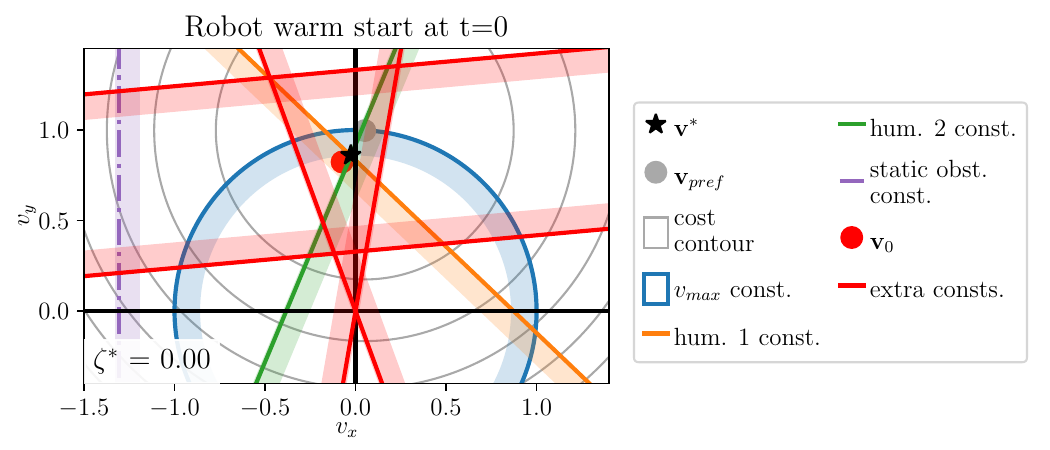}
    \caption{Illustration of the optimization problem used to solve for robot actions for each step of the warm start for the same environment state as Fig.~\ref{fig:env_setup_posns}. The robot's kino-dynamic constraints are underestimated by the red linear constraints. The extra constraints are identically formulated as the static obstacle constraints in \eqref{eq:relaxed_orca_argmin}.}
	\label{fig:warmstart}
\end{figure}

We introduce a warm start strategy in order to initialize the optimization of the \ac{MPC} problem \eqref{eq:combmpckkt_prob} from a feasible trajectory and to mitigate problems caused by scenarios where the lower-level problems do not meet the \ac{LICQ}. To obtain a feasible trajectory for all agents, we initialize the optimization with trajectories generated from rollouts of the \ac{ORCA} policy for the robot in addition to the humans.
Using \ac{ORCA} for the robot will allow us to satisfy its safety constraints. In order to also satisfy the robot's kino-dynamic constraints, we alter the \ac{ORCA} optimization used for the robot.
We illustrate the altered \ac{ORCA} policy in Fig.~\ref{fig:warmstart}. Unlike the humans which are modelled as holonomic integrators, the robot has non-holonomic kino-dynamic constraints \eqref{eq:combmpckkt_rob_dynconst}-\eqref{eq:combmpckkt_actionconst} (unicycle model and limits change in velocity) described in section~\ref{sec:sysstatedyn}. To account for these constraints, we add four extra pseudo-collision constraints, illustrated in red in Fig.~\ref{fig:warmstart}, the first two (crossing lines) bound the change in heading of the robot, and the second two (parallel lines) bound the change in linear velocity of the robot. Since \ac{ORCA} constraints are linear in the velocity state, we set the constraints to be an underestimate the robot's actual constraints, hence ensuring that the rollout is feasible, albeit with more conservative constraints.

We use the results of a sequential rollout of this optimization problem as the initial guess for the robot's actions for the first time we solve the \ac{MPC} problem.
For subsequent \ac{MPC} optimizations, we bring forward the \ac{MPC} solution from the previous time step by shifting the trajectory by one time step, and use the altered \ac{ORCA} policy to update the remaining (final) time step of the \ac{MPC} horizon.
In the case that for some attempted \ac{MPC} solution we are unable to converge to a trajectory that is feasible and lower-cost than the initial guess, e.g. when we encounter a configuration where one of the lower-level problems does not satisfy \ac{LICQ}, we use the warm start solution instead.
\begin{table*}[th!]
    \centering
    \caption{Comparison of trajectory prediction methods on (a) \acs{ADE}/\acs{FDE}  of the \textit{maximum likelihood} prediction, (b) \acs{ADE}/\acs{FDE} of the \textit{best-of-20} samples, and (c) mean \acs{KDE} \acs{NLL}. {\color{blue} Our \ac{ORCA}-based open-loop predictions do not achieve state-of-the-art performance but are nonetheless competitive with top-performing methods.}}
    \begin{tabular}{|l||c|c|c|c||c|c|c|c||c|c|c|}
    \hline
    \multirow{2}{*}{\textbf{Dataset}} & \multicolumn{4}{c||}{\rule{0pt}{2ex} \textbf{(a)} Maximum likelihood \acs{ADE}/\acs{FDE} (m) $\downarrow$}& \multicolumn{4}{c||}{\textbf{(b)} Best-of-20 \acs{ADE}/\acs{FDE} (m) $\downarrow$} & \multicolumn{3}{c|}{\textbf{(c)} Mean KDE NLL $\downarrow$} \label{tab:orcapred}\\
    \cline{2-12}
                     & \rule{0pt}{2.5ex} CVMM      & S-GAN \cite{gupta2018sgan}      & Traj++  \cite{saltzmann2021trajpp}    & $\mathcal{O}$ (ours) & Y-Net \cite{mangalam2021ynet}      & S-GAN       & Traj++      & $\mathcal{O}$ (ours) & S-GAN   & Traj++  & $\mathcal{O}$ (ours) \\ \hline
    ETH              & 1.15/3.92                   & {1.09/2.34} & {0.71/1.66} & 0.97/1.85            & {0.28/0.33} & {0.71/1.36} & {0.39/0.83} & {0.59/1.00}          & 4.42  & {1.31}  &  4.76                \\
    Hotel            & 0.23/0.74                   & {0.69/1.44} & {0.22/0.46} & 0.28/0.51            & {0.10/0.14} & {0.37/0.72} & {0.12/0.21} & {0.13/0.21}          & 4.22 & {-1.29} & -0.20                \\
    Univ             & 0.71/0.99                   & {0.71/1.55} & {0.44/1.17} & 0.64/1.33            & {0.24/0.41} & {0.55/1.16} & {0.20/0.44} & {0.33/0.64}          & 5.30 & {-0.89} &  1.92                \\
    Zara 1           & 0.58/1.08                   & {0.64/1.43} & {0.30/0.79} & 0.51/1.07            & {0.17/0.27} & {0.32/0.65} & {0.15/0.33} & {0.25/0.48}          & 4.20 & {-1.13} &  0.62                \\
    Zara 2           & 0.43/0.34                   & {0.52/1.14} & {0.23/0.59} & 0.39/0.82            & {0.13/0.22} & {0.30/0.63} & {0.11/0.25} & {0.21/0.40}          & 3.62 & {-2.19} &  0.45                \\
    \hline
    \end{tabular}
    \label{tab:comparison}
\end{table*}
\section{Evaluating ORCA for trajectory forecasting} \label{sec:pred}
In this paper we argue that our proposed method, \ac{SICNav}, has the properties of being \textit{interactive} in that the robot is able to explicitly model and optimize its influence on the other agents, and \textit{safe} in that the resulting \ac{MPC} policy satisfies explicit collision constraints. %
However, for these properties to hold in the real world we need to demonstrate that the \ac{ORCA} policy is a sufficiently accurate model for predicting real human trajectories. The authors in \cite{Chen2022orcampc} have previously shown that \ac{ORCA} can be used to generate accurate predictions of humans in multi-agent scenarios. In this section, we follow a similar approach to validate the quality of stand-alone trajectory predictions produced by running rollouts of the \ac{ORCA} algorithm for several time steps on a human trajectory dataset.

In order to generate predictions using a rollout of \ac{ORCA}, we require a goal-location for each pedestrian, which we use to determine the preferred velocity for the \ac{ORCA} optimization cost.
The method in \cite{Chen2022orcampc} uses a naive-Bayesian classifier to select a goal for each agent from a finite set of potential goal locations, then, performs a rollout to generate a single deterministic prediction per agent. {\color{blue} However, a priori knowledge of such a set of potential goals is not available in many crowd navigation environments. Furthermore}, most state-of-the-art trajectory forecasting methods (e.g. \cite{saltzmann2021trajpp, mangalam2021ynet,yue2022nspsfm}), additionally model the uncertainty of predictions in order to account for the multi-modality of human motion. Thus, in order to fairly compare to these methods, we introduce a method to sample goal positions from a training dataset of human trajectories, then produce predictions with \ac{ORCA} for each sampled goal. Unlike \cite{Chen2022orcampc}, our method does not require a finite set of a priori known goal positions, we rather sample from the distribution generated from the training split of the dataset. We compare our method with state-of-the-art trajectory forecasting methods on the ETH \cite{pellegrini2009ethdata} and UCY \cite{lerner2007ucydata} pedestrian datasets, a commonly used benchmark for trajectory forecasting.

\subsection{Methodology}
Given a history of states (positions and velocities) with length $T_h = 3.2s$, $\id{\idA}{\at{\x}{-T_h:0}}$, for a pedestrian, we want to predict a sequence of future positions with length $T_p = 4.8s$, $\id{\idA}{\at{\p}{0:T_p}}$. In order to use roll-outs of the \ac{ORCA} policy, we first need to predict a goal position for each agent. To this end, we use the trajectories in the training dataset to generate goal distributions to sample. The training set contains full trajectories for each agent, $\id{\idA}{\at{\x}{-T_h:T_p}}$. First, we create a 2D histogram of the trajectories in the training set based on average historic velocity and acceleration. In each bin, we obtain the goal location as the final position of the agent at the end of each trajectory in the bin, $\id{\idA}{\at{\p}{T_p}}$. For all the goal positions in each bin, we fit a 2D \ac{KDE} to estimate the density of goal positions.

At test time, we have a scene with $\numhumans$ pedestrians. For each agent in the scene, we use its average historic velocity to select the appropriate goal $\ac{KDE}$, and obtain $2000$ samples of goal positions for each agent. This way, we are drawing independent goal samples for each agent. However, in order to perform a rollout of the scene using \ac{ORCA}, we need to perform a rollout for the entire scene, i.e. we need samples from the joint-distribution of goals for every agent. If we naively combine the goals of each agent to get $2000$ joint samples, we ignore the fact that a highly likely goal sample for one agent may get matched with a low likely goal for another. To address this issue, we calculate importance weights for the combined goal samples. We sum the log-likelihoods of each individual goal sample over the number of agents in the scene and normalize with respect to the $2000$ samples. Now, for each of these $2000$ joint samples, we perform an \ac{ORCA} rollout.

\subsection{Results}
We follow the \textit{leave-one-out} strategy commonly used with the ETH/UCY datasets \cite{saltzmann2021trajpp, mangalam2021ynet}, where one dataset is selected as the test set, and the others are used for training (generating the goal \acp{KDE} in our formulation).
Following the metrics used in existing literature \cite{saltzmann2021trajpp, mangalam2021ynet}, we evaluate prediction using,
\begin{enumerate}
    \item \ac{ADE}: the mean Euclidean distance between the ground truth and the maximum likelihood prediction.
    \item \ac{FDE}: the Euclidean distance between the maximum likelihood predicted position at the final prediction step, $T_p$, and the ground truth.
    \item Best-of-20: the minimum \ac{ADE} and minimum \ac{FDE} from 20 randomly sampled prediction trajectories.
    \item \ac{KDE}-based \ac{NLL} \cite{Thiede2019KDE}: the mean \ac{NLL} of the ground truth trajectory under the distribution of a \ac{KDE} fit to the predicted trajectory samples.
\end{enumerate}
To find the maximum likelihood prediction from our model, we follow a similar approach as calculating the \ac{KDE}-based \ac{NLL} metric, however, instead of evaluating the likelihood of the ground truth under the \acp{KDE}, we evaluate the likelihoods of the predicted samples themselves.
We compare our method against two state-of-the-art prediction models Social-GAN \cite{gupta2018sgan}, Trajectron++ \cite{saltzmann2021trajpp} and Y-net \cite{mangalam2021ynet}. For the maximum likelihood metrics, we also compare our method to a \ac{CVMM} that simply projects the current velocity of the agent forward in time. We were unable to replicate the results of Y-net, and thus cannot report the KDE-NLL metric.

The results of our evaluation are summarized in Table~\ref{tab:orcapred}.
We can observe that our \ac{ORCA}-based approach does not achieve state-of-the-art performance but is nonetheless competitive with top-performing methods. Our method beats S-GAN and CVMM on almost all metrics for all data splits. Our method has the added benefit of being able to be integrated into downstream planning, which is challenging for the other methods.

\section{Evaluation of SICNav}
\subsection{Simulation Experiments} \label{sec:sim_results}
\emph{Testing environment}: We evaluate the performance of our planner in a commonly used crowd navigation simulation environment \cite{Chen2019c}, which we extend to simulate static obstacles {\color{blue} in addition to the original dynamic agents, and to enable the simulated agents to be controlled by \ac{SFM} \cite{Helbing1995sfm} in addition to the original \ac{ORCA}}.
{\color{blue} We conduct two sets of experiments, one where the simulated humans follow \ac{ORCA} and another where they follow \ac{SFM}. For each set, we generate 500 random initial and final positions for the humans environments with $\numhumans=3$ and $\numhumans=5$ simulated humans to get a total of 2000 runs.
The simulated testing area is} a $2m$-wide corridor with initial positions and goals on either end. In order to ensure a high density of agents in the corridor, we constrain the space using static obstacles, formulated as line-segments, and add a bottleneck that simulates a $1m$ wide doorway. Fig.~\ref{fig:interactive_snapshots2} highlights an interaction at this doorway. In every scenario, the robot is required to move through the simulated doorway to a goal $3m$ directly in front its fixed initial position.
Furthermore, every agent must cross the doorway in order to arrive at its goal, ensuring that interactions occur in these scenarios.

\emph{Performance metrics}: We adopt several metrics from the literature \cite{Chen2019c, Kretzschmar2016, Trautman2010}. As the robot proceeds to the goal, data from the trajectories of the robot and agents is collected to score the behavior in the following ways,
\begin{itemize} 
	\item \textit{Success Rate:} The rate of scenarios in which the robot is able to successfully arrive at its goal within $90s$, which measures how the overall success of the robot to arrive at its goal.
	\item \textit{Average Navigation Time:} The average time to completion, which indicates the robot's time efficiency.
	\item \textit{Collision Frequency:} The ratio of time steps that the robot is in a collision state with another agent compared to the total number of time-steps in the 500 scenarios, which measures the safety of the robot.
	\item \textit{Freezing Frequency:} The frequency of time steps where the robot needed to reduce its velocity to zero, which measures how often the robot can potentially get stuck in a deadlock with the human agents (aka the encountering the \ac{FRP}).
\end{itemize}
\begin{table}[t]
    \caption{Quantitative comparison in 500 randomly generated test scenarios with 1 robot and 3 or 5 simulated humans {\color{blue} following \ac{ORCA}}. Arrows indicate desired relative values. {\color{blue} Our method outperforms baselines accross the metrics.}}  \label{tab:sim_results}
    \begin{center}
        \begin{tabular}{|l|c|c|c|c|} 
            \hline
            \multicolumn{1}{|p{1.2cm}|}{\centering \vfill \textbf{Approach}} & \multicolumn{1}{p{1.0cm}|}{\centering \textbf{Success Rate} $\uparrow$} &  \multicolumn{1}{p{1.3cm}|}{\centering \textbf{Avg Nav Time (s)} $\downarrow$} & \multicolumn{1}{p{1.0cm}|}{\centering \textbf{Collision Freq.} $\downarrow$} & \multicolumn{1}{p{1.0cm}|}{\centering \textbf{Frozen Freq.} $\downarrow$} \\
            \hline
            \multicolumn{5}{|c|}{\rule{0pt}{2ex} $\numhumans=3$ humans} \\
            \hline
            SICNav-p (ours)                     & $\mathbf{1.00}$ 			   & $\mathbf{4.24}$
																												& $\mathbf{0.00}$ & $\mathbf{0.01}$\\
            \color{blue} SICNav-np (ours)       & \color{blue} $\mathbf{1.00}$ & \color{blue} $4.37$
																												& \color{blue} $\mathbf{0.00}$ & $\mathbf{0.01}$\\
            MPC-CVMM                            & $0.98$          			   & $7.06$
																												& ${0.01}$        & ${0.05}$ \\
            \ac{ORCA} \cite{vandenBerg2011orca} & $\mathbf{1.00}$ 			   & $10.26$
																												& ${0.01}$        & ${0.07}$ \\
            SARL \cite{Chen2019c}               & $0.73$          			   & $27.21$
																												& ${0.01}$        & ${0.26}$ \\
            RGL  \cite{chen_relational_2020}    & $0.95$          			   & $8.20$
																												& ${0.01}$        & ${0.06}$ \\
            DWA  \cite{Fox1997dwa}              & $0.57$          			   & $52.09$
																												& ${0.01}$        & ${0.55}$ \\ \hline
            \multicolumn{5}{|c|}{\rule{0pt}{2ex} $\numhumans=5$ humans} \\
            \hline
            SICNav-p (ours)                     & $\mathbf{1.00}$ 			   & $\mathbf{6.35}$
																											& $\mathbf{0.01}$ & $\mathbf{0.03}$\\
            \color{blue} SICNav-np (ours)       & \color{blue} $\mathbf{1.00}$ & \color{blue} $\mathbf{6.27}$
																											& \color{blue} $\mathbf{0.01}$ & \color{blue} $\mathbf{0.02}$\\
            MPC-CVMM                            & $\mathbf{1.00}$ 			   & $7.47$
																											& ${0.05}$        & ${0.07}$ \\
            \ac{ORCA} \cite{vandenBerg2011orca} & $\mathbf{1.00}$ 			   & $14.98$
																											& ${0.02}$        & ${0.09}$ \\
            SARL \cite{Chen2019c}               & $0.97$          			   & ${6.45}$
																											& ${0.05}$        & ${0.04}$ \\
            RGL  \cite{chen_relational_2020}    & $0.50$          			   & $46.86$
																											& ${0.03}$        & ${0.48}$ \\
            DWA  \cite{Fox1997dwa}              & $0.68$          			   & $48.45$
																											& ${0.03}$        & ${0.50}$ \\
            \hline
        \end{tabular}
    \end{center}
\end{table}
\begin{table}[t]
    \color{blue}
    \caption{\color{blue} Quantitative comparison in 500 randomly generated test scenarios with 1 robot and 3 or 5 simulated humans following \ac{SFM}. Arrows indicate desired relative values. Compared with performance in the ORCA simulations, our method does not show much degradation in any of the metrics.}  \label{tab:sim_results_sfm}
    \begin{center}
        \begin{tabular}{|l|c|c|c|c|} 
            \hline
            \multicolumn{1}{|p{1.2cm}|}{\centering \vfill \textbf{Approach}} & \multicolumn{1}{p{1.0cm}|}{\centering \textbf{Success Rate} $\uparrow$} &  \multicolumn{1}{p{1.3cm}|}{\centering \textbf{Avg Nav Time (s)} $\downarrow$} & \multicolumn{1}{p{1.0cm}|}{\centering \textbf{Collision Freq.} $\downarrow$} & \multicolumn{1}{p{1.0cm}|}{\centering \textbf{Frozen Freq.} $\downarrow$} \\
            \hline
            \multicolumn{5}{|c|}{\rule{0pt}{2ex} $\numhumans=3$ humans} \\
            \hline
            SICNav-p (ours)                     & $\mathbf{1.00}$ & $4.84$
																						& $0.01$          & $\mathbf{0.01}$\\
            SICNav-np (ours)                    & $\mathbf{1.00}$ & $\mathbf{4.81}$
																						& $0.01$          & $\mathbf{0.01}$\\
            MPC-CVMM                            & $0.99$          & $6.25$
																						& ${0.02}$        & ${0.03}$ \\
            \ac{ORCA} \cite{vandenBerg2011orca} & $\mathbf{1.00}$ & $6.10$
																						& $0.01$          & $\mathbf{0.01}$ \\
            SARL \cite{Chen2019c}               & $0.64$          & $34.82$
																						& ${0.02}$        & ${0.36}$ \\
            RGL  \cite{chen_relational_2020}    & $0.95$          & ${7.49}$
																						& ${0.02}$        & ${0.05}$ \\
            DWA  \cite{Fox1997dwa}              & $0.96$          & $16.66$
																						& $\mathbf{0.00}$ & ${0.10}$ \\ \hline
            \multicolumn{5}{|c|}{\rule{0pt}{2ex} $\numhumans=5$ humans} \\
            \hline
            SICNav-p (ours)                     & $\mathbf{1.00}$ & $5.42$
																						& ${0.03}$        & $\mathbf{0.02}$\\
            SICNav-np (ours)                    & $\mathbf{1.00}$ & $\mathbf{5.14}$
																						& ${0.02}$        & $\mathbf{0.01}$\\
            MPC-CVMM                            & $0.98$          & $7.30$
																						& ${0.05}$        & ${0.05}$ \\
            \ac{ORCA} \cite{vandenBerg2011orca} & $\mathbf{1.00}$ & $7.83$
																						& ${0.03}$        & ${0.02}$ \\
            SARL \cite{Chen2019c}               & $0.92$          & ${10.27}$
																						& ${0.03}$        & ${0.08}$ \\
            RGL  \cite{chen_relational_2020}    & $0.17$          & $75.56$
																						& $\mathbf{0.01}$ & ${0.80}$ \\
            DWA  \cite{Fox1997dwa}              & $0.95$          & $20.78$
																						& $\mathbf{0.01}$ & ${0.14}$ \\
            \hline
        \end{tabular}
    \end{center}
\end{table}
\begin{figure*}[th]
	\footnotesize
	\begin{subfigure}[]{\textwidth}
	    \footnotesize
		\includegraphics[width=\textwidth]{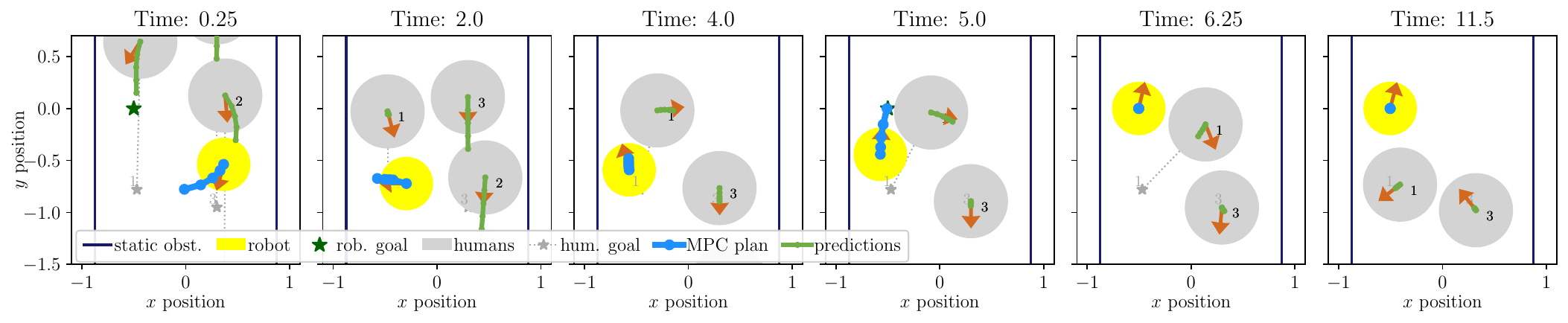}
		\caption{SICNav-p (ours)}
	    \label{fig:sicnav_qualitative}
	\end{subfigure}
	\begin{subfigure}[]{\textwidth}
	    \footnotesize
		\includegraphics[width=\textwidth]{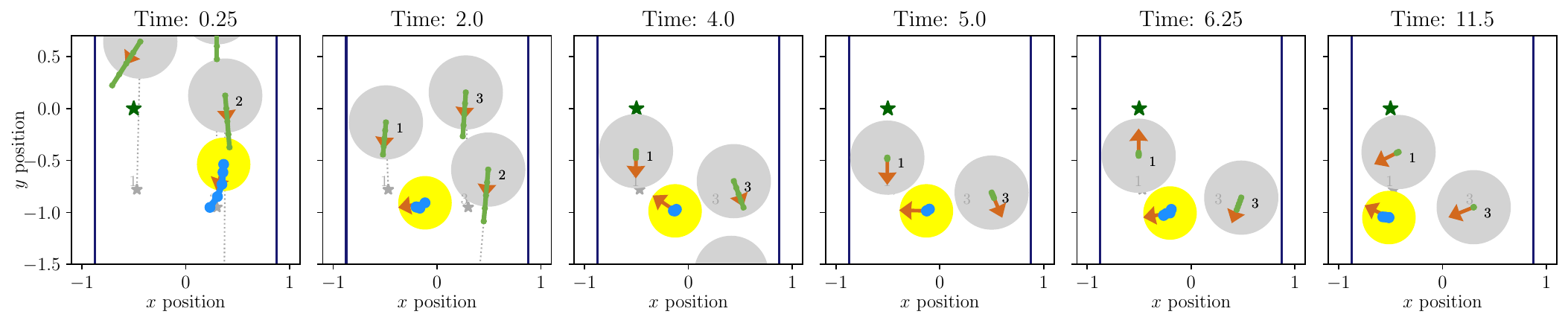}
		\caption{MPC-CVMM}
	    \label{fig:mpccvmm_qualitative}
	\end{subfigure}
    \caption{Snapshots through time of a robot (yellow) using our SICNav approach (a) and the MPC-CVMM baseline approach (b), near a goal location (green star) obstructed by simulated human agents (gray circles). At time 0.25, the robot and the humans are moving downwards, and both the SICNav and MPC-CVMM robot plan to turn to the right towards their goal (MPC plan in blue). Our SICNav robot arrives at the goal by time 6.25 by predicting that human 1 will move out of its way (predictions in green), then proceed to its own goal, arriving at time 11.5. Meanwhile, the robot using the baseline MPC-CVMM method is unable to coordinate with human 1. Animation: \href{https://tiny.cc/sicnav_fig5}{\texttt{tiny.cc/sicnav\_fig5}}.}
    \label{fig:interactive_snapshots}
\end{figure*}

{\emph{Algorithms}: \color{blue}We compare two variants of our algorithm with four baseline methods. The first version of our algorithm, SICNav-p, has access to the ground truth goal of position and \ac{ORCA} parameters of the human agents. The second version, SICNav-np, does not have access to any privileged information about the agents. In order to estimate the intended goal of the agent, SICNav-np projects the current velocity of the agent forward in time, and uses the resulting position as the goal of the human. We compare these two versions of our algorithm with the following baselines:
}
MPC-CVMM has the same setup as SICNav, however with a \ac{CVMM} instead of the lower-level \ac{ORCA} model for other agents, i.e. each human is assumed to continue at their current velocity for the complete horizon.
We also compare with state of the art \ac{RL} methods, \ac{SARL} \cite{Chen2019c}, and \ac{RGL} \cite{chen_relational_2020}, the robot itself also running \ac{ORCA} \cite{vandenBerg2011orca}, and the classical \ac{DWA} \cite{Fox1997dwa} {\color{blue} as a reactive baseline}. Note that since \ac{ORCA} does not have a method to incorporate dynamics constraints by default, the \ac{ORCA} robot follows a holonomic model without any kino-dynamic (unicycle) constraints, which means that it solves a simpler problem than the other methods.

\emph{Implementation}\footnote{Code: \href{https://github.com/sepsamavi/safe-interactive-crowdnav.git}{\scriptsize \texttt{github.com/sepsamavi/safe-interactive-crowdnav.git}}.}: We implement SICNav in Python using CasADi \cite{Andersson2019casadi}. To avoid infeasibility we introduce slack variables on the constraints along with quadratic penalty functions. We use a simulation time step of $\delta t = 0.25s$ and an \ac{MPC} horizon of $\horiz = 1s$. The \ac{RL} methods are implemented using Stable Baselines \cite{stable-baselines}. The cost functions for both \ac{RL} methods have been altered to include penalties analogous to all components of SICNav's cost and constraints described in Sec.~\ref{sec:our_soln}, and the one-step \ac{CVMM} propagation used to calculate the respective value functions has been replaced with \ac{ORCA} in order to allow fairer comparisons to our SICNav method, which uses the \ac{ORCA} model in the lower-level problem. We train the \ac{RL} algorithms with the same arrangement of static obstacles as the test environment, but with a training set of human initial and goal positions that are not included in the test set. We train separate models for $\numhumans=3$ and $\numhumans=5$. We intend to make the code for the simulator and methods publicly available upon acceptance.

\begin{figure*}[th]
	\centering
	\footnotesize
	\begin{subfigure}[]{\textwidth}
		\centering
	    \footnotesize
		\includegraphics[width=\textwidth]{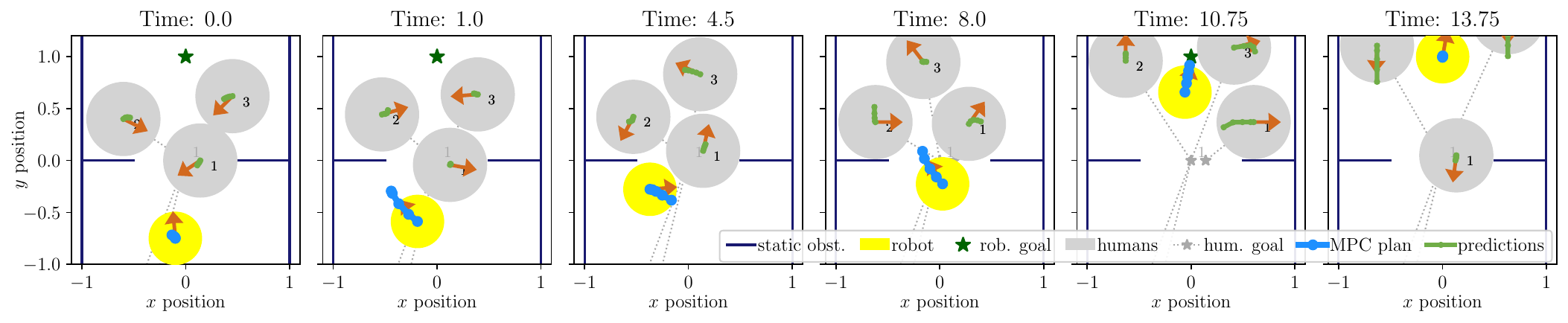}
		\caption{SICNav-p (planning with \ac{ORCA} predictions computed jointly)}
	    \label{fig:sicnav_qualitative2}
	\end{subfigure}
	\hfill
	\begin{subfigure}[]{\textwidth}
		\centering
	    \footnotesize
		\includegraphics[width=\textwidth]{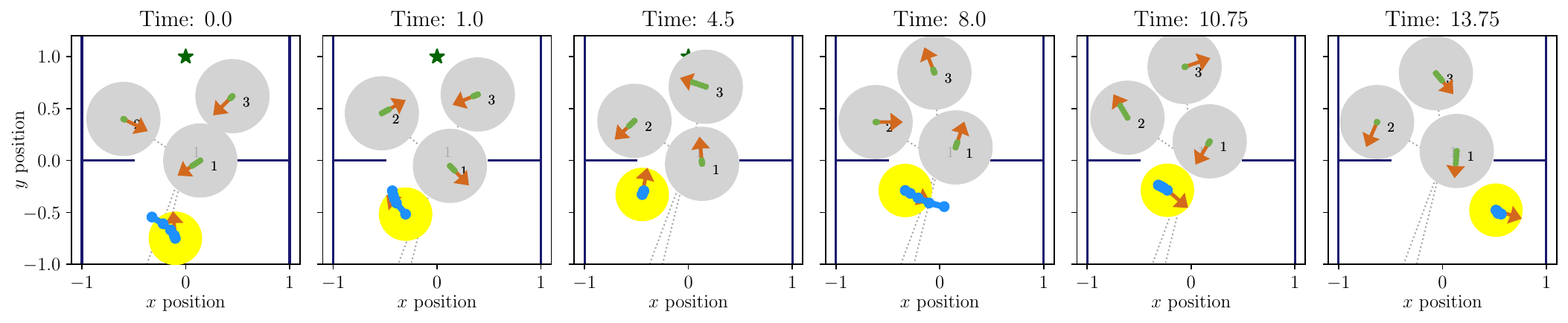}
		\caption{MPC-CVMM (planning with CVMM predictions that remain static)}
	    \label{fig:mpccvmm_qualitative2}
	\end{subfigure}
    \caption{Snapshots through time of a robot (yellow) using our SICNav-p approach (a) and the MPC-CVMM baseline approach (b), near a goal location (green star) obstructed by simulated human agents (gray circles). The goal position of human 1 is in the way of the robot's path to its goal. At time 0.0, the robot begins moving towards the doorway. As shown in the images, our SICNav-p robot continues moving to its goal by predicting that agents 1-3 will move out of its way. However, in the same scenario, the robot using the baseline MPC-CVMM method oscillates between positions on the left and right of agent 1, who remains blocking the doorway. Animation: \href{https://tiny.cc/sicnav_fig6}{\texttt{tiny.cc/sicnav\_fig6}}.}
    \label{fig:interactive_snapshots2}
\end{figure*}
\emph{Quantitative Results and Discussion}:
{\color{blue}
Tables~\ref{tab:sim_results}~and~\ref{tab:sim_results_sfm} summarize the performance of our method and the baselines in the simulator with \ac{ORCA} agents and \ac{SFM} agents, respectively. In the following section we analyze the average performance of the crowd navigation methods. A statistical analysis of the significance of these results is provided in Appendix~\ref{app:stat_sim_results}.}

In simulation scenarios involving three humans {\color{blue} simulated by \ac{ORCA}}, the two variants our method, SICNav-p and {\color{blue}SICNav-np}, achieve a success rate of 1.00, beating all other methods, except \ac{ORCA} which achieves the same success rate. Accompanying SICNav-p's success rate are the shortest average navigation time (4.24s) and the lowest rates of collision (0.00) and robot freezing (0.01). {\color{blue} Removing privileged information results in negligibly slower navigation time for SICNav-np (4.37), without affecting collision and freezing rates.} Using the \ac{ORCA} for robot plans, however, demonstrates a much longer navigation time in excess of 10s, and a collision rate of 0.01. This poorer performance compared to SICNav-p is despite the fact that the ORCA method is solving a simpler problem without kino-dynamic constraints. RGL demonstrates an average navigation time of 8.20s and success rate of 0.95. The MPC-CVMM method achieves a slightly higher success rate at 0.98, but requires a longer average navigation time (7.06s). SARL, experiences a lowered success rate (0.73), a high average navigation time (27.21s), and increased rates of collision (0.01) and robot freezing (0.26). DWA records the lowest success rate (0.57), the longest average navigation time (52.09s), and a high rate of robot freezing (0.55).

In the scenarios with five humans {\color{blue} simulated by \ac{ORCA}}, all SICNav-p, SICNav-np, MPC-CVMM, and \ac{ORCA} maintain a success rate of 1.00.
{\color{blue} Similar to the three human tests, SICNav-p and SICNav-np perform better than all the baselines across all four metrics, indicating our method's consistent performance under increasing crowd densities. Unlike the three human tests, SICNav-np posts the shortest average navigation time at 6.27s compared to SICNav-p's 6.35s, with both versions of our method maintaining the lowest rates for collision (0.01) and robot freezing (0.02, and 0.03, respectively)}. MPC-CVMM and ORCA record longer average navigation times (7.47s and 14.98s) and show substantially higher rates of collisions (0.05 and 0.02) and robot freezing (0.05 and 0.09). In comparison, SARL reached an average navigation time of 6.45s, but with a success rate of 0.97 and also with a substantial rate of collisions (0.05). RGL and DWA, under the same conditions, demonstrate lower success rates (0.50 and 0.68 respectively), extended navigation times (46.86s and 48.45s) and increased rates of collisions and robot freezing.

{\color{blue}
In simulation scenarios involving three humans simulated by \ac{SFM}, the two variants of our method, SICNav-p and SICNav-np, achieve a perfect success rate of 1.00, maintaining consistent performance with the different human simulation model. SICNav-np, slightly outperforms SICNav-p in terms of average navigation time, recording 4.81s compared to 4.84s with similar collision frequencies of 0.01. The baseline methods exhibit varied performance. For instance, \ac{ORCA} shows improved navigation time (6.10s) for this human simulation model but with an equivalent collision frequency of 0.01. The MPC-CVMM method achieves a success rate slightly lower at 0.99 and a longer average navigation time (6.25s). SARL significantly struggles with the lowest success rate (0.64) and a much higher average navigation time (34.82s), along with increased collision and freezing rates, suggesting a severe degradation of performance when faced with novel behavior from the simulated humans. With this human simulation model, DWA outperforms our methods in terms of collision frequency (0.00) however at the expense of a lower success rate (0.96), a lengthier navigation times (16.66s), and increased freezing frequency (0.10).

In scenarios with five humans simulated by \ac{SFM}, both of our models (SICNav-p and SICNav-np) continue to exhibit perfect success rates. SICNav-np presents shorter navigation times (5.14s), slightly better than SICNav-p's 5.42s. The collision rates for both our models remain low but slightly increased to 0.02 and 0.03, respectively, indicating our model being strained under higher densities of \ac{SFM} humans. MPC-CVMM and \ac{ORCA} demonstrate longer navigation times (7.30s and 7.83s) and higher rates of both collisions and freezing. SARL's success rate depletes to 0.92 with an inflated navigation time of 10.27s. RGL's performance notably deteriorates, culminating in a drastically reduced success rate of 0.17 and an extended average navigation time. The degradation in performance of the \ac{RL} methods in the \ac{SFM} simulator demonstrates the challenge these methods have when faced with human behavior that deviates from the training data. DWA, however, maintains a high success rate of 0.95 and a low collision frequency of 0.01, but with a longer navigation time (20.78s) and increased freezing frequency (0.14).

The results in with the \ac{SFM} simulated humans are consistent with the results from the \ac{ORCA} simulated humans. Our proposed methods, SICNav-p and SICNav-np, do not show much degradation in any of the metrics, demonstrating the robustness of our approach to different human simulation models.

}
{\color{blue}
\begin{figure*}[th!]
	\centering
	\footnotesize
	\begin{subfigure}[]{0.15\textwidth}
		\centering
	    \footnotesize
		\includegraphics[width=\textwidth]{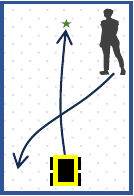}
		\caption{$\numhumans=1$}
	    \label{fig:1hum_env}
	\end{subfigure}
	\hfill
    \begin{subfigure}[]{0.15\textwidth}
		\centering
	    \footnotesize
		\includegraphics[width=\textwidth]{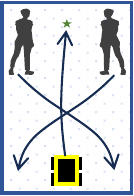}
		\caption{$\numhumans=2$ (variant~1)}
	    \label{fig:1hum_env}
	\end{subfigure}
    \hfill
    \begin{subfigure}[]{0.15\textwidth}
		\centering
	    \footnotesize
		\includegraphics[width=\textwidth]{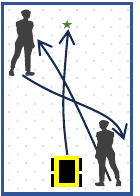}
		\caption{$\numhumans=2$ (variant~2)}
	    \label{fig:1hum_env}
	\end{subfigure}
	\hfill
    \begin{subfigure}[]{0.15\textwidth}
		\centering
	    \footnotesize
		\includegraphics[width=\textwidth]{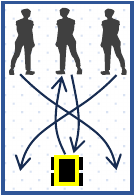}
		\caption{$\numhumans=3$ (variant~1)}
	    \label{fig:1hum_env}
	\end{subfigure}
    \hfill
    \begin{subfigure}[]{0.15\textwidth}
		\centering
	    \footnotesize
		\includegraphics[width=\textwidth]{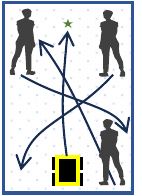}
		\caption{$\numhumans=3$ (variant~2)}
	    \label{fig:1hum_env}
	\end{subfigure}
    \caption{{\color{blue} Types of scenarios that constitute the 80 runs of the robot tested with humans (not to scale). The robot (black and yellow) follows a straight path to the goal (green star). The humans start from the positions shown in the figure and move to the opposite side of the testing area as shown by the solid black arrows. The agents begin moving at the same time. All scenarios were designed to induce a conflict zone in the middle of the room where all agents' paths cross. Video of scenarios: \href{https://tiny.cc/sicnav_robot_scenarios}{\texttt{tiny.cc/sicnav\_robot\_scenarios}}. }}
	\label{fig:exp_scenarios}
\end{figure*}
}
\emph{Qualitative Analysis}: To qualitatively demonstrate the impact of SICNav-p generating robot plans jointly with predictions, in Fig.~\ref{fig:interactive_snapshots}, we illustrate a scenario in which the robot cannot access its goal unless it interacts with other agents blocking its way. Since our SICNav-p method can proactively expect the reaction of the human agents, it slowly approaches the agents in its way, who move to allow the robot to pass in response. Note that the robot is not colliding with the other agents, the agents move out of its way and allow it to pass. On the same scenario we compare the baseline MPC-CVMM, whose cost and constraints are identical to SICNav-p except for the predictions of the other agents which use a \ac{CVMM} model and remain constant in the planning optimization. Starting from an almost identical position, MPC-CVMM gets stuck as it is unable to account for the cooperation of the other agents.

Fig.~\ref{fig:interactive_snapshots2} illustrates a similar but more complex scenario where the robot cannot access its goal unless it interacts with other agents blocking a doorway. In this scenario, the SICNav-p robot is able to move through the doorway by predicting that the agents will move out of its way. However, the baseline MPC-CVMM robot oscillates back and forth around without proceeding through the doorway.

{\color{blue}
\subsection{Real Robot Experiments} \label{sec:robot_results}
\emph{Testing Environment}: We also evaluate the performance of our planner, SICNav-np, on a real robot in a controlled environment. We use a Clearpath Jackal differential drive robot pictured in Fig.~\ref{fig:problem_overview}, which weighs $20 kg$ and measures $46cm$ W $\times 60cm$ L $\times \times 50cm$ H. Our operating environment is an indoor space measuring $5m \times 9m$. We conduct a total of 80 runs from five scenarios with a varying number of humans $\numhumans \in \set{1,2,3}$. The scenarios are illustrated in Fig.~\ref{fig:exp_scenarios}. For each run of a particular scenario, all agents were instructed to move from their starting positions (where the agents are illustrated in Fig.~\ref{fig:exp_scenarios}) to the opposite corner of the room (illustrated by the arrows in Fig.~\ref{fig:exp_scenarios}). In the scenarios with multiple humans, we tested two variants: one where all the humans start from the opposite side of the room than the robot (variant 1) and another where one human moved in the same direction as the robot (variant 2). These scenarios were designed to cause a conflict zone in the middle of the room where all agents need to interact. We repeated 20 runs for each scenario with $\numhumans \in \set{1,2}$ humans and 10 runs for each scenario with $\numhumans = 3$.
We use a VICON motion capture system to measure the positions of the robot and human agents in the environment and use an \ac{EKF} and \acp{KF} to track the positions and velocities of the robot and humans, respectively.

\begin{figure*}[th]
	\centering
	\footnotesize
	\begin{subfigure}[]{0.49\textwidth}
		\centering
	    \footnotesize
		\includegraphics[width=\textwidth]{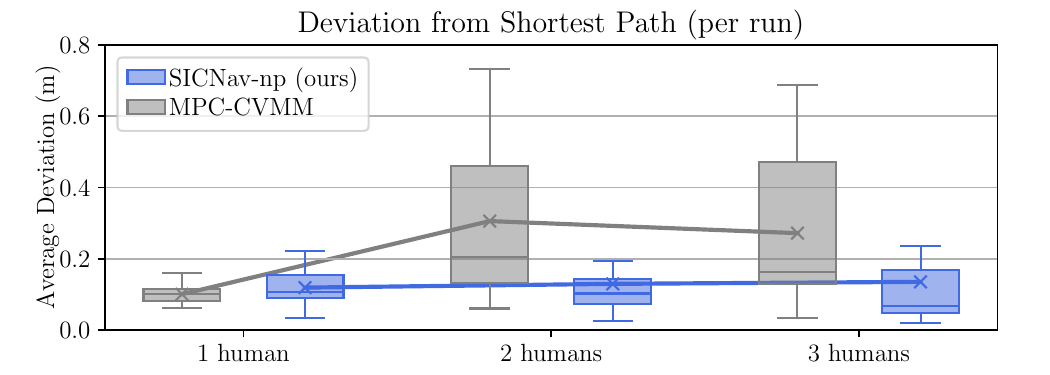}
		\caption{Deviation from the shortest path}
	    \label{fig:real_robot_deviation}
	\end{subfigure}
	\hfill
    \begin{subfigure}[]{0.49\textwidth}
		\centering
	    \footnotesize
		\includegraphics[width=\textwidth]{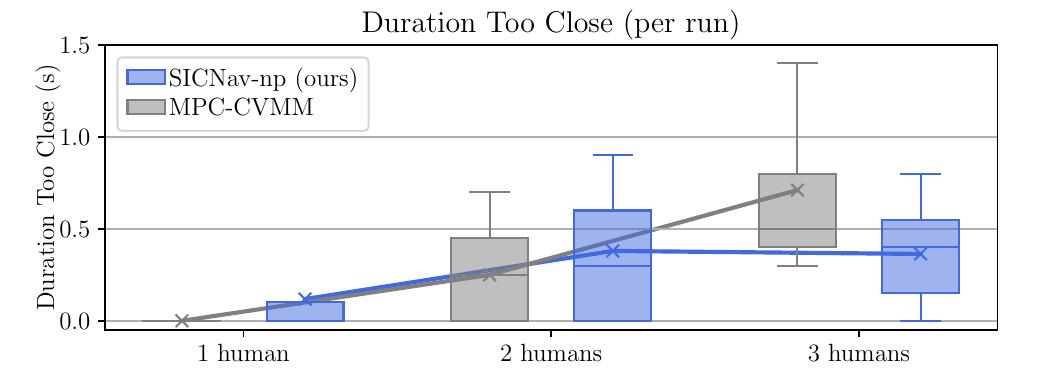}
		\caption{Duration too close to a human}
	    \label{fig:real_robot_tooclose}
	\end{subfigure}
    \caption{{\color{blue} Box-and-whisker plots of performance metrics obtained when running SICNav-np (blue) and MPC-CVMM (gray) in scenarios with varying numbers of humans. Median values are illustrated by the line in the box-and-whisker plots and mean values are indicated by the $\times$. We plot (a) the robot's average deviation from the shortest path in each run, and (b) the total duration of time that the robot is too close to any human in each run. As crowd size increases, using our method instead of the baseline results in a lower deviation from the shortest path, demonstrating better path efficiency, and a lower duration too close to a human, demonstrating better safety.}}
    \label{fig:real_robot_results}
\end{figure*}
\begin{figure*}[th]
	\centering
	\footnotesize
	\begin{subfigure}[]{0.49\textwidth}
		\centering
	    \footnotesize
		\includegraphics[width=\textwidth]{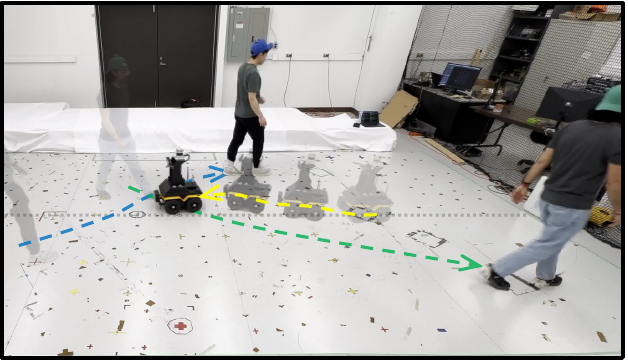}
		\caption{SICNav-np (ours)}
	    \label{fig:sicnav_qualitative2}
	\end{subfigure}
	\hfill
	\begin{subfigure}[]{0.49\textwidth}
		\centering
	    \footnotesize
		\includegraphics[width=\textwidth]{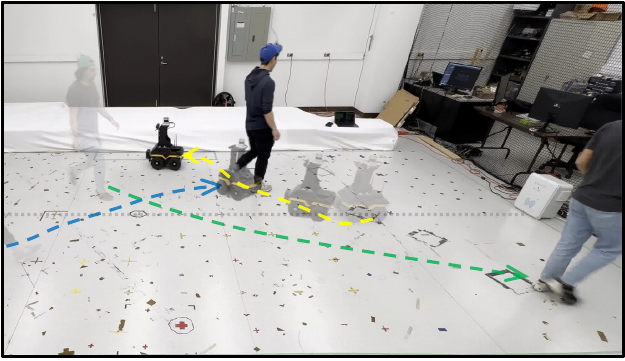}
		\caption{MPC-CVMM (baseline)}
	    \label{fig:mpccvmm_qualitative2}
	\end{subfigure}
    \caption{{\color{blue} Videos of one robot and two humans rendered as images. Dashed lines indicate paths traversed by each agent (yellow for robot, green and blue for humans). Transparent images illustrate older frames of the video. For both SICNav-np (a) and the baseline (b), the robot begins moving towards its goal in the center of the room (gray dotted line). Our SICNav robot's path (yellow dashed line in (a)) deviates less from the center of the room (gray dotted line) than the path of the robot using the baseline MPC-CVMM method (yellow dashed line in (b)). Video: \href{https://tiny.cc/sicnav_fig9}{\texttt{tiny.cc/sicnav\_fig9}}.}} \label{fig:interactive_snapshots_real1}
\end{figure*}
\begin{figure*}[th]
	\centering
	\footnotesize
	\begin{subfigure}[]{0.49\textwidth}
		\centering
	    \footnotesize
		\includegraphics[width=\textwidth]{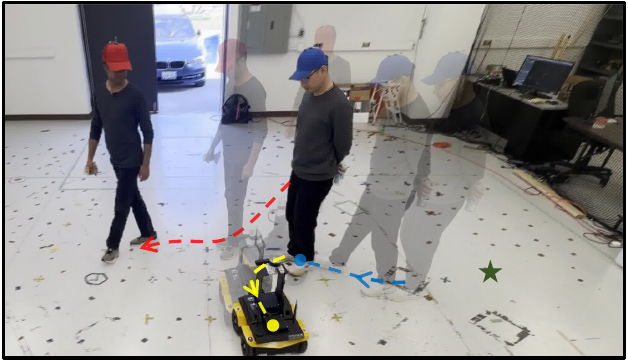}
		\caption{\texttt{00:03} to \texttt{00:11} SICNav-np robot being pushed back by humans}
	    \label{fig:adverse_real_pushback}
	\end{subfigure}
	\hfill
	\begin{subfigure}[]{0.49\textwidth}
		\centering
	    \footnotesize
		\includegraphics[width=\textwidth]{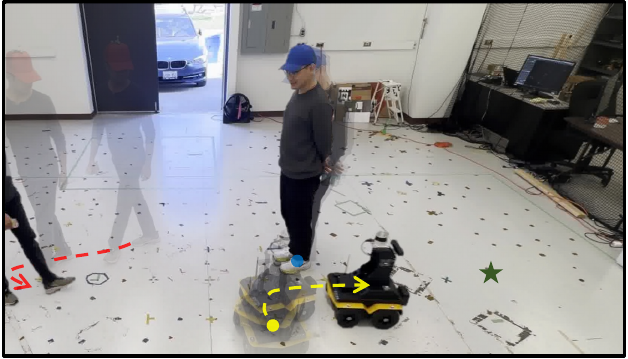}
		\caption{\texttt{00:11} to \texttt{00:18} SICNav-np robot moving forward after being pushed}
	    \label{fig:adverse_real_fwd}
	\end{subfigure}
	\begin{subfigure}[]{\textwidth}
	    \footnotesize
		\includegraphics[width=\textwidth]{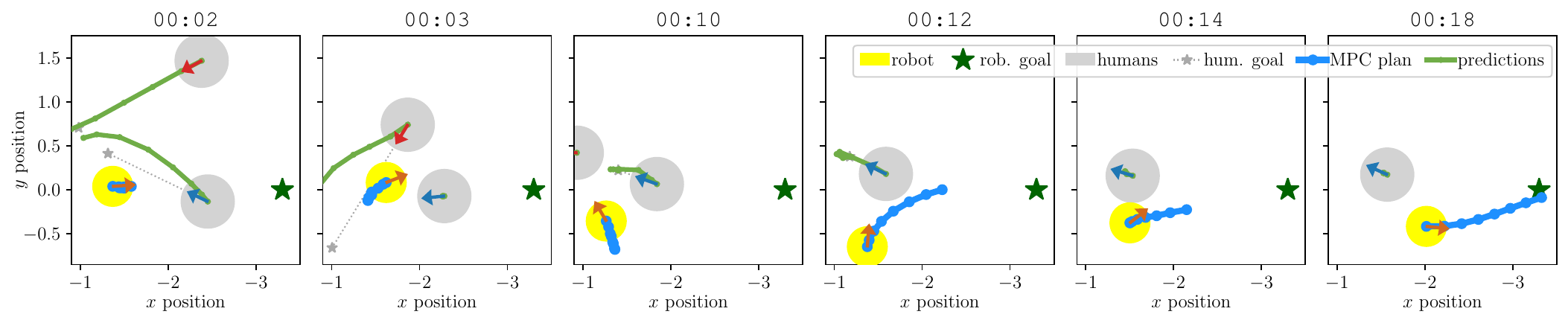}
		\caption{Top view snapshots of SICNav-np robot being pushed back and going forward}
	    \label{fig:adverse_real_snapshots}
	\end{subfigure}
    \caption{{\color{blue} Videos of one robot and two humans rendered as images. Dashed lines indicate paths traversed by each agent (yellow for robot, blue and red for humans). Transparent images illustrate older frames of the video. The robot goal is illustrated as the green star. (a) As the robot proceeds towards its goal, the humans with the blue and red hats walk towards the robot without attempting to avoid it, resulting in the robot being pushed back. (b) After the humans stop pushing, the robot comes to a stop and moves forward around the human with the blue hat to proceed to its goal. In (c), we illustrate snapshots of the top-view visualization corresponding to (a) and (b). Video: \href{https://tiny.cc/sicnav_fig10}{\texttt{tiny.cc/sicnav\_fig10}}.}} \label{fig:adverse_real}
\end{figure*}

\emph{Algorithms}: We compare the performance our SICNav-np method with MPC-CVMM as a baseline. As noted in Sec.~\ref{sec:sim_results}, SICNav-np, does not have access to any privileged information about the agents. In order to estimate the intended goal of the agent, SICNav-np projects the current velocity of the agent forward in time, and uses the resulting position as the human's goal. MPC-CVMM has the same setup as SICNav-np, however with a \ac{CVMM} instead of the lower-level \ac{ORCA} model for other agents, i.e. each human is assumed to continue at their current velocity for each step of the \ac{MPC} horizon. By comparing our method with MPC-CVMM, we aim to isolate our analysis specifically to the primary contribution of this paper: the effect of using \ac{ORCA} to model the humans and jointly produce robot plans with predictions.

\emph{Implementation}: The robot used the \ac{ROS} 1 middleware. The robot accepts linear and angular velocity commands, which are then executed by its low level controller. We run the algorithms under analysis on an 8-core computer using Intel$^{(R)}$ Core$^{(TM)}$ i7-9700K CPU @ 3.6 GHz. We use the same Python implementation of our algorithm as the simulation experiments (Sec.~\ref{sec:sim_results}), which uses CasADi as the modelling language. However, we replace the solver with Acados \cite{verschueren2021acados}, which generates C++ code from the CasADi models and uses Just-in-time Compilation to allow us to run our algorithm in real-time at 10 Hz. We keep the same discretization step as the simulation of $\delta t = 0.25s$ and use an \ac{MPC} horizon of $\horiz = 2s$.

\emph{Quantitative Results and Discussion}: %
We evaluate the robot's path efficiency by measuring the average distance of the robot from the shortest path to its goal in each run. In Fig.~\ref{fig:real_robot_deviation}, we observe that in scenarios with $\numhumans \in \set{2,3}$, SICNav-np has a lower average deviation from the shortest path, with $0.13m$ and $0.17m$, compared to MPC-CVMM, with $0.30m$ and $0.27m$. Furthermore, we observe that the distribution of values for MPC-CVMM is larger than for SICNav-np. These results indicate that SICNav-np is more efficient in terms of the path it follows, and more consistently chooses the shorter path.

We evaluate the robot's safety by measuring the total duration of time the robot is too close to any human (body of robot is $< 0.35 m$ to centroid of human) in each run. In Fig.~\ref{fig:real_robot_tooclose} we observe that in the scenarios with $\numhumans \in \set{1,2}$ there is not much difference in the average amount of time the robot is too close to a human between the two algorithms, with an average total time of $0.12s$ and $0.38s$ for SICNav-np compared to $0.00s$ and $0.26s$ for MPC-CVMM. However, in the case with $\numhumans = 3$ humans, we can see that SICNav-np has a lower average duration too close to a human, with $0.36s$, compared to $0.77s$ for MPC-CVMM. While SICNav-np and MPC-CVMM perform similarly with respect to scenarios with two or fewer humans, in scenarios with a higher number of humans, SICNav-np maintains roughly the same amount of duration too close to any human, while the time for MPC-CVMM almost doubles, indicating that SICNav-np is more successful in avoiding close interactions with humans. We can also observe that while the size of the interquartile range is similar between the two methods for cases with $\numhumans \in \set{1,2}$, in the cases with $\numhumans=3$, the interquartile range is smaller for SICNav-np, indicating that SICNav-np is more consistent in avoiding close interactions with humans.
\emph{Qualitative Analysis}: %
Reviewing the videos of the runs, we observe that the robot using SICNav-np would often slow down and wait for the human to cross its path before proceeding. We illustrate this observation in Fig.~\ref{fig:interactive_snapshots_real1} (Video: \href{https://tiny.cc/sicnav_fig9}{\texttt{tiny.cc/sicnav\_fig9}}). The SICNav-np robot predicts that the humans will alter their paths to pass each other and with the robot. As such the robot slows down. In contrast, the robot using MPC-CVMM, predicts that the humans would continue at their current speed and heading. Thus, the MPC-CVMM robot turns to the right to avoid the human's path leading to a larger deviation from the shortest path than the SICNav-np robot. The observations from this scenario are consistent with the results in Fig.~\ref{fig:real_robot_deviation}, where we observe that SICNav-np has a lower average deviation from the shortest path than MPC-CVMM.

In Fig.~\ref{fig:adverse_real}, we illustrate a scenario where the robot is confronted by humans acting adversarially towards the robot (Video: \href{https://tiny.cc/sicnav_fig10}{\texttt{tiny.cc/sicnav\_fig10}}). %
Fig.~\ref{fig:adverse_real_pushback} illustrates the beginning of this scenario: as the robot proceeds towards its goal (green star), the humans with the blue and red hats walk towards the robot without attempting to avoid it. Snapshots of the top-view are illustrated in the first three plots in Fig.~\ref{fig:adverse_real_snapshots}, starting with the humans walking toward the robot at time \texttt{00:02}.  At times \texttt{00:02} and \texttt{00:03}, we observe that the predictions produced by SICNav-np (green lines) expect the humans to go around the robot. In the associated robot MPC plans (blue lines) we can see the robot planning to reverse away from the humans in the first few time steps of the MPC horizon, then proceed forward in the latter part of the horizon. However, in this scenario, the humans do not follow the robot's predictions that they will deviate their path to avoid the robot, instead they proceed toward the robot. This human behavior deviates from our assumption that the humans follow \ac{ORCA}. However, since the robot follows SICNav-np in receding horizon fashion, re-planning at 10Hz, new plans take into account the fact that the humans are getting closer to the robot than the SICNav-np prediction expected. As such, subsequent robot plans for the robot to reverse at the beginning of the MPC horizon in order to continue avoiding the humans. Thus, we observe the robot being pushed back, as illustrated in Fig.~\ref{fig:adverse_real_pushback}. Once the humans stop behaving adversarially, at time \texttt{00:11} in this scenario, the robot predictions become more accurate (e.g. human with the blue hat will remain roughly stationary at \texttt{00:14}), and the robot proceeds to its goal as illustrated in Fig.~\ref{fig:adverse_real_fwd}. In this scenario, we observe the robustness of our method to adversarial behavior from the humans. In such scenarios, the robot only needs to avoid collisions for one time step of the MPC horizon. If its predictions are incorrect, as long as the robot can avoid the humans for one time step, it can re-plan and continue avoiding collisions.
}
\subsection{Limitations and Future Work} \label{sec:limitations_future_work}
Our evaluation of SICNav in simulation {\color{blue} and with a real robot operating in a lab setting} show promising results. {\color{blue} Our algorithm runs in real time and enables the robot to} interact with other agents while avoiding collisions in tight scenarios with multiple agents.
Nonetheless, our paper has several limitations that we intend to address in future work.

{\color{blue}
\emph{Knowledge of Human Intents}: SICNav requires information about the human agents' intended goals in order to predict their future trajectory. In SICNav-np, which does not require knowledge of human goals, we project each agent's current velocity forward in time to estimate their goal. While this method shows promising performance in our real-robot experiments (Sec.~\ref{sec:robot_results}),} in future work, we plan to use a learning-based short-term goal prediction (e.g. \cite{mangalam2021ynet}) to generate goal estimates that we can then use in the \ac{ORCA} models integrated into SICNav. To train, we intend to combine a goal predictor with open-loop \ac{ORCA} rollouts and use a standard prediction loss (e.g. \cite{saltzmann2021trajpp,mangalam2021ynet}), leveraging the implicit function theorem to differentiate the \ac{ORCA} $\argmin\{\cdot\}$ as a function. We can even extend this approach to infer the other \ac{ORCA} parameters from data as well. Our evaluation of \ac{ORCA} rollouts as an open-loop prediction model (Sec.~\ref{sec:pred}) has shown promising results, even with a rudimentary empirical goal sampling method and \ac{ORCA} parameters fixed. We also intend to investigate incorporating uncertainty of the predicted \ac{ORCA} parameters into the SICNav bilevel optimization, e.g. by following a robust bilevel optimization approach \cite{beck2021robustbilevel}.

{\color{blue}
\emph{Evaluation Scope}: In this work, we evaluate SICNav in simulation and on a real robot operating indoors. In our robot experiments, we use a VICON motion tracking system to obtain position measurements (mm error) of the humans head position in the scene, and use \acp{KF} to estimate their positions and velocities. The first limitation of this setup is that we would not have access to such motion tracking infrastructure in a deployment in the wild. Nonetheless, recent human motion estimation work (e.g.\cite{xie2021physicsbased}) has shown promising results in estimating and predicting human motion from only image data to centimeter accuracy. Furthermore, our estimates of the human centroid did not meet the mm accuracy possible VICON: a second limitation of our experimental setup is that we only tracked the location of each agent's head. Thus, our estimates of the person's location were noisy when the person swayed or leaned, e.g. to initiate or stop walking. Previous work using images and lidar \cite{burnett2019autotrack}, has shown position tracking \ac{RMSE} of less than 15cm for nearby agents. Toward the key next step of deploying SICNav on a real robot in the wild, we intend to evaluate our algorithm's performance when using the onboard lidar and camera sensors to track the positions of the agents in the scene.

\emph{Number of Interactive Humans}: In this work, we demonstrate the real-time (10Hz) performance of our algorithm in environments with up to $\numhumans=3$ humans modeled using \ac{ORCA} for a horizon of $\horiz=8$ time steps.} The dimension of variables for the \ac{MPC} optimization problem \eqref{eq:combmpckkt_prob} associated with the \ac{KKT}-reformulation scale quadratically in the number of humans times the control horizon, $O(\numhumans^2\horiz)$. As a result, solving for control may become slower than real time for scenarios with many agents {\color{blue}($>> 3$)}. In future work, we intend to limit the number of agents modeled interactively {\color{blue} using ORCA} to a neighborhood of the {\color{blue}$3-4$} most `important' humans {\color{blue} each time we solve \eqref{eq:combmpckkt_prob} and use the solution in receding horizon fashion}, similar to \cite{Trautman2015,DuToit2012}.
\section{Conclusion} \label{sec:conclusion}
In this paper we presented \ac{SICNav}, a novel \ac{MPC} planner for robot navigation in crowded environments. We formulated the problem as a bilevel optimization problem, where the robot plans its trajectory while jointly predicting the trajectories of other agents using \ac{ORCA}. To evaluate our approach, we first validated the effectiveness of \ac{ORCA} in generating open-loop predictions of human motion on a benchmark dataset. Then, we demonstrated the effectiveness of our planning method in simulation {\color{blue} environments with \ac{ORCA}-simulated humans and \ac{SFM}-simulated humans}, and showed that our method outperforms state-of-the-art methods in terms of success rate, navigation time, and collision avoidance. We also qualitatively evaluated the ability of our method to interact with other agents in tight simulation scenarios. {\color{blue} Finally, we evaluated our method on a real robot in a lab environment, demonstrating our methods improved navigation performance when compared to a baseline. We also qualitatively evaluated our method in cases where the humans deviate from behaving as expected under \ac{ORCA} to demonstrate our method's robustness.} %
{\color{blue}
\appendix
\subsection{Statistical Analysis of Simulation Results} \label{app:stat_sim_results}

In this section, we perform a series of statistical tests to verify that the crowd navigation simulation experiments presented in Section~\ref{sec:sim_results} show statistically significant improvements in performance by using our proposed method.

\subsubsection{Selection of Statistical Tests}
In order to determine whether to use parametric or non-parametric tests, we first perform Shapiro-Wilk tests to check for normality of the data for each metric (Navigation Time, Collision Frequency, and Frozen Frequency) that is measured over the 500 test scenarios for each algorithm (SICNav, SICNav-np, MPC-CVMM, ORCA, SARL, RGL, DWA) in each environment ($\numhumans=3$ humans, $\numhumans=5$ humans). We conduct a total of 42 Shapiro-Wilk tests (3 metrics $\times$ 7 algorithms $\times$ 2 environments).
The hypotheses in each test are,
\begin{itemize}
    \item $H_0$: The performance metrics  normally distributed.
    \item $H_1$: The data is not normally distributed.
\end{itemize}
We use a significance level of $\alpha=0.05$ for all tests. 

The results of the Shapiro-Wilk test are summarized in Tables~\ref{tab:shapiro_wilk}~and~\ref{tab:shapiro_wilk_sfm}.
\begin{table}[t]
    \color{blue}
    \caption{\color{blue}Shapiro-Wilk Test for Normality across Metrics and Algorithms tested in \ac{ORCA} simulations. Arrows indicate desired values ($\uparrow$ indicates desired higher p-value toward rejecting $H_0$, i.e. accepting normality).}  \label{tab:shapiro_wilk}
    \begin{center}
        \begin{tabular}{|l|c|c|c|} %
            \hline
            \textbf{Approach} & \multicolumn{1}{p{1.5cm}|}{\centering \textbf{Navigation Time p-value} $\uparrow$} & \multicolumn{1}{p{1.5cm}|}{\centering\textbf{Collision Frequency p-value} $\uparrow$} & \multicolumn{1}{p{1.5cm}|}{\centering\textbf{Frozen Freqency p-value} $\uparrow$ }\\
            \hline
            \multicolumn{4}{|c|}{\rule{0pt}{2ex} $\numhumans=3$ humans} \\
            \hline
            SICNav-p        & $5.59e-43$ & $2.35e-42$ & $3.25e-41$ \\
            SICNav-np       & $3.53e-31$ & $1.02e-40$ & $1.53e-39$ \\
            MPC-CVMM        & $7.74e-41$ & $4.94e-39$ & $1.82e-38$ \\
            ORCA            & $1.00$     & $1.00$     & $1.00$ \\
            DWA             & $1.00$     & $1.00$     & $1.00$ \\
            SARL            & $2.57e-33$ & $1.16e-39$ & $6.38e-33$ \\
            RGL             & $1.57e-40$ & $4.07e-38$ & $1.09e-39$ \\
            \hline
            \multicolumn{4}{|c|}{\rule{0pt}{2ex} $\numhumans=5$ humans} \\
            \hline
            SICNav-p        & $1.44e-32$ & $4.20e-40$ & $4.22e-38$ \\
            SICNav-np       & $3.95e-28$ & $1.17e-35$ & $3.46e-37$ \\
            MPC-CVMM        & $4.52e-28$ & $5.49e-31$ & $2.26e-31$ \\
            ORCA            & $1.00$     & $1.00$     & $1.00$ \\
            DWA             & $1.00$     & $1.00$     & $1.00$ \\
            SARL            & $2.21e-41$ & $1.96e-33$ & $5.46e-41$ \\
            RGL             & $6.97e-31$ & $1.08e-34$ & $3.64e-30$ \\
            \hline
        \end{tabular}
    \end{center}
\end{table}
\begin{table}[t]
    \color{blue}
    \caption{\color{blue}Shapiro-Wilk Test for Normality across Metrics and Algorithms tested in \ac{SFM} simulations. Arrows indicate desired values ($\uparrow$ indicates desired higher p-value toward rejecting $H_0$, i.e. accepting normality).}  \label{tab:shapiro_wilk_sfm}
    \begin{center}
        \begin{tabular}{|l|c|c|c|} %
            \hline
            \textbf{Approach} & \multicolumn{1}{p{1.5cm}|}{\centering \textbf{Navigation Time p-value} $\uparrow$} & \multicolumn{1}{p{1.5cm}|}{\centering\textbf{Collision Frequency p-value} $\uparrow$} & \multicolumn{1}{p{1.5cm}|}{\centering\textbf{Frozen Frequency p-value} $\uparrow$ }\\
            \hline
            \multicolumn{4}{|c|}{\rule{0pt}{2ex} $\numhumans=3$ humans} \\
            \hline
            SICNav-p        & $9.30e-43$  & $2.08e-34$  & $4.42e-41$  \\
            SICNav-np       & $1.31e-41$  & $4.00e-35$  & $1.37e-41$  \\
            MPC-CVMM        & $4.28e-42$  & $9.83e-37$  & $4.27e-40$  \\
            ORCA            & $3.04e-18$  & $7.35e-38$  & $7.60e-37$  \\
            DWA             & $6.71e-33$  & $1.03e-43$  & $5.72e-33$  \\
            SARL            & $1.11e-31$  & $4.97e-40$  & $8.27e-30$  \\
            RGL             & $5.14e-41$  & $1.13e-38$  & $2.98e-40$  \\
            \hline
            \multicolumn{4}{|c|}{\rule{0pt}{2ex} $\numhumans=5$ humans} \\
            \hline
            SICNav-p        & $1.31e-29$  & $4.79e-36$  & $2.05e-37$  \\
            SICNav-np       & $1.94e-28$  & $2.90e-35$  & $2.88e-38$  \\
            MPC-CVMM        & $2.62e-41$  & $1.15e-32$  & $4.39e-39$  \\
            ORCA            & $1.87e-35$  & $1.70e-31$  & $2.58e-38$  \\
            DWA             & $2.42e-30$  & $5.85e-43$  & $3.05e-30$  \\
            SARL            & $2.14e-39$  & $1.62e-37$  & $5.83e-39$  \\
            RGL             & $2.77e-36$  & $9.66e-41$  & $2.24e-35$  \\
            \hline
        \end{tabular}
    \end{center}
\end{table}
We observe that the p-values for the Shapiro-Wilk tests are all far below $\alpha$ for all algorithms and metrics, except for all metrics in ORCA and DWA in the \ac{ORCA} simulation environments. Since these results indicate that the performance results are not normally distributed for many algorithms, we need to use non-parametric tests.

\subsubsection{Mann-Whitney U Test for Pair-Wise Comparisons}
We use the Mann-Whitney U test to evaluate the significance of the differences observed in the performance metrics evaluated on the 500 evaluation scenarios across pairs of algorithms in each environment.

We begin by comparing the two variants of our method, SICNav and SICNav-np. We conduct a total of $6$ Mann-Whitney U tests (3 metric $\times$ 2 environments) for this pair of algorithms.
The hypotheses in each test are,
\begin{itemize}
    \item $H_0$: The distributions of the performance metrics are identical between SICNav and SICNav-np.
    \item $H_1$: The distributions of the performance metrics are not identical between SICNav and SICNav-np.
\end{itemize}
As before, we use a significance level of $\alpha=0.05$.
Tables~\ref{tab:pairwise_sicnav_vs_sicnav_np}~and~\ref{tab:pairwise_sicnav_vs_sicnav_np_sfm} summarize the results of the Mann-Whitney U tests comparing SICNav-p and SICNav-np in the \ac{ORCA} and \ac{SFM} simulations, respectively. We observe that the test statistics for all metrics and environments are high,
\begin{table}[t]
    \color{blue}
    \caption{\color{blue}Pairwise Mann-Whitney U Test Results between SICNav-p and SICNav-np tested in \ac{ORCA} simulations. Arrows indicate the desired outcome toward failing to reject the null hypothesis, signifying no significant differences between SICNav-p and SICNav-np.}
    \label{tab:pairwise_sicnav_vs_sicnav_np}
    \begin{center}
        \begin{tabular}{|l|c|c|c|c|} %
        \hline
        & \multicolumn{2}{c|}{$N = 3$ humans} & \multicolumn{2}{c|}{$N = 5$ humans} \\
        \cline{2-5}
        \textbf{Metric} & \textbf{Stat.} $\uparrow$ & \textbf{p-value} $\uparrow$ & \textbf{Stat.} $\uparrow$ & \textbf{p-value} $\uparrow$ \\
        \hline\hline
        Nav. Time       & 121221.5 & 0.402296 & 119372.5 & 0.216947 \\
        Coll. Freq. & 120554.0 & 0.009947 & 110152.0 & $1.30e-07$ \\
        Frozen Freq.    & 124767.0 & 0.919510 & 123475.0 & 0.609077 \\
        \hline
        \end{tabular}
    \end{center}
\end{table}
\begin{table}[t]
    \color{blue}
    \caption{\color{blue}Pairwise Mann-Whitney U Test Results between SICNav-p and SICNav-np tested in \ac{SFM} simulations. Arrows indicate the desired outcome toward failing to reject the null hypothesis, signifying no significant differences between SICNav-p and SICNav-np.}
    \label{tab:pairwise_sicnav_vs_sicnav_np_sfm}
    \begin{center}
        \begin{tabular}{|l|c|c|c|c|} %
        \hline
        & \multicolumn{2}{c|}{$N = 3$ humans} & \multicolumn{2}{c|}{$N = 5$ humans} \\
        \cline{2-5}
        \textbf{Metric} & \textbf{Stat.} $\uparrow$ & \textbf{p-value} $\uparrow$ & \textbf{Stat.} $\uparrow$ & \textbf{p-value} $\uparrow$ \\
        \hline\hline
        Nav. Time       & 117642.5 & 0.104038 & 133368.0 & 0.065717 \\
        Coll. Freq.     & 126088.5 & 0.741520 & 128207.0 & 0.380315 \\
        Frozen Freq.    & 126975.5 & 0.422128 & 132179.0 & 0.013285 \\
        \hline
        \end{tabular}
    \end{center}
\end{table}

From the pairwise tests in the SFM environment, we observe from the results that for navigation time and frozen frequency, the p-values exceed the significance level. This indicates no significant difference in the distributions of these performance metrics between SICNav-p and SICNav-np in both environments (i.e. failure to reject $H_0$). Conversely, for collision frequency, the observed p-values ($0.009947$ in the $N = 3$ humans environment and $1.30e-07$ in the $N = 5$ humans environment) fall below the significance threshold. This suggests that the distributions of collision frequencies between SICNav-p and SICNav-np are significantly different (i.e. reject $H_0$).

From the pairwise tests in the SFM environment, we also observe that for navigation time and frozen frequency, the p-values ($0.104038$ and $0.422128$ for $N = 3$ humans; $0.065717$ and $0.013285$ for $N = 5$ humans) generally exceed the significance level, except for the frozen frequency in the $N = 5$ humans scenario, indicating no significant difference in these performance metrics between SICNav-p and SICNav-np in most cases. In contrast, for collision frequency, the p-values ($0.741520$ for $N = 3$ humans and $0.380315$ for $N = 5$ humans) exceed the significance level in the SFM environment, further indicating no significant difference for collision frequency in this environment.

These tests provide compelling evidence that the absence of privileged information in SICNav-np does not significantly affect the navigation time and frozen frequency in both ORCA and SFM environments, but does significantly affect the collision frequency in the ORCA environment. However, the average difference in collision frequency between SICNav-p and SICNav-np in the ORCA environment are small (see Table~\ref{tab:sim_results}). Nonetheless, for the sake of completeness, we proceed to compare both SICNav-p and SICNav-np against each of the baseline algorithms.

For each of the two variants of our proposed method, SICNav-p and SICNav-np, we conduct a total of $30$ Mann-Whitney U tests (3 metrics $\times$ 5 baseline algorithms $\times$ 2 environments). The hypotheses in each test are,
\begin{itemize}
    \item $H_0$: The distributions of the performance metrics are identical between our proposed algorithm and the baseline algorithm.
    \item $H_1$: The distributions of the performance metrics are not identical between our proposed algorithm and the baseline algorithm.
\end{itemize}
As before, we use a significance level of $\alpha=0.05$.

\begin{table}[t]
    \color{blue}
    \caption{\color{blue}Pairwise Mann-Whitney U Test Results of SICNav-p against Baseline Algorithms tested in \ac{ORCA} simulations. Arrows in the table ($\uparrow$ for Stat. and $\downarrow$ for p-value) represent our analytical desire to uncover significant differences between SICNav-p and each baseline, indicating SICNav-p's distinct performance advantages.}
    \label{tab:pairwise_sicnav_vs_baselines}
    \begin{center}
        \begin{tabular}{|l|c|c|c|c|} %
        \hline
        & \multicolumn{2}{c|}{$N = 3$ humans} & \multicolumn{2}{c|}{$N = 5$ humans} \\
        \cline{2-5}
        \textbf{Metric} & \textbf{Stat.} $\uparrow$ & \textbf{p-value} $\downarrow$ & \textbf{Stat.} $\uparrow$ & \textbf{p-value} $\downarrow$ \\
        \hline\hline
         \multicolumn{5}{|c|}{\rule{0pt}{2ex} {SICNav-p vs. MPC-CVMM}} \\
        \hline
        Nav. Time       & 85827.0 & $4.06e-18$ & 108493.0 & $0.000292915$ \\
        Coll. Freq. & 111530.0 & $1.31e-09$ & 85984.0 & $2.58e-30$ \\
        Frozen Freq.    & 110833.5 & $4.07e-07$ & 103446.0 & $3.14e-10$ \\
        \hline
         \multicolumn{5}{|c|}{\rule{0pt}{2ex} {SICNav-p vs. ORCA}} \\
        \hline
        Nav. Time       & 40075.0 & $3.48e-77$ & 26466.5 & $6.54e-103$ \\
        Coll. Freq. & 112631.5 & $3.10e-08$ & 100827.5 & $1.50e-14$ \\
        Frozen Freq.    & 84343.0 & $3.06e-31$ & 79141.0 & $4.70e-32$ \\
        \hline
        \multicolumn{5}{|c|}{\rule{0pt}{2ex} {SICNav-p vs. DWA}} \\
        \hline
        Nav. Time       & 4656.0 & $5.80e-155$ & 10950.5 & $1.53e-137$ \\
        Coll. Freq. & 115431.5 & $3.88e-06$ & 71260.5 & $5.47e-46$ \\
        Frozen Freq.    & 24000.0 & $1.03e-127$ & 18861.5 & $4.25e-132$ \\
        \hline
        \multicolumn{5}{|c|}{\rule{0pt}{2ex} {SICNav-p vs. SARL}} \\
        \hline
        Nav. Time       & 164736.0 & $1.42e-18$ & 215377.0 & $5.50e-88$ \\
        Coll. Freq. & 109332.5 & $1.54e-11$ & 93854.5 & $4.88e-22$ \\
        Frozen Freq.    & 95514.0 & $1.43e-20$ & 128810.5 & $0.176$ \\
        \hline
        \multicolumn{5}{|c|}{\rule{0pt}{2ex} {SICNav-p vs. RGL}} \\
        \hline
        Nav. Time       & 179211.5 & $4.60e-33$ & 107574.0 & $0.000119$ \\
        Coll. Freq. & 109442.0 & $1.71e-11$ & 75747.5 & $6.22e-41$ \\
        Frozen Freq.    & 116993.5 & $0.002$ & 66160.0 & $4.91e-51$ \\
        \hline
        \end{tabular}
    \end{center}
\end{table}
\begin{table}[t]
    \color{blue}
    \caption{\color{blue}Pairwise Mann-Whitney U Test Results of SICNav-p against Baseline Algorithms tested in \ac{SFM} simulations. Arrows in the table ($\uparrow$ for Stat. and $\downarrow$ for p-value) represent our analytical desire to uncover significant differences between SICNav-p and each baseline, indicating SICNav-p's distinct performance advantages.}
    \label{tab:pairwise_sicnav_vs_baselines_sfm}
    \begin{center}
        \begin{tabular}{|l|c|c|c|c|} %
        \hline
        & \multicolumn{2}{c|}{$N = 3$ humans} & \multicolumn{2}{c|}{$N = 5$ humans} \\
        \cline{2-5}
        \textbf{Metric} & \textbf{Stat.} $\uparrow$ & \textbf{p-value} $\downarrow$ & \textbf{Stat.} $\uparrow$ & \textbf{p-value} $\downarrow$ \\
        \hline\hline
        \multicolumn{5}{|c|}{\rule{0pt}{2ex} {SICNav-p vs. MPC-CVMM}} \\
        \hline
        Nav. Time       & 100911.5 & $1.10e-07$ & 129735.0 & $0.298018$ \\
        Coll. Freq.     & 123682.5 & $0.691238$ & 113669.5 & $0.003175$ \\
        Frozen Freq.    & 112400.0 & $2.06e-05$ & 117566.5 & $0.022995$ \\
        \hline
        \multicolumn{5}{|c|}{\rule{0pt}{2ex} {SICNav-p vs. ORCA}} \\
        \hline
        Nav. Time       & 59444.5  & $4.82e-47$ & 58916.0  & $1.39e-47$ \\
        Coll. Freq.     & 136718.5 & $0.000103$ & 120294.5 & $0.216113$ \\
        Frozen Freq.    & 120162.0 & $0.077268$ & 111429.5 & $0.000104$ \\
        \hline
        \multicolumn{5}{|c|}{\rule{0pt}{2ex} {SICNav-p vs. DWA}} \\
        \hline
        Nav. Time       & 6776.5   & $3.91e-148$& 12364.5  & $1.80e-134$ \\
        Coll. Freq.     & 145570.0 & $1.16e-13$ & 153497.5 & $1.10e-19$ \\
        Frozen Freq.    & 72485.5  & $9.05e-44$ & 71533.0  & $1.43e-40$ \\
        \hline
        \multicolumn{5}{|c|}{\rule{0pt}{2ex} {SICNav-p vs. SARL}} \\
        \hline
        Nav. Time       & 140532.5 & $0.000622$ & 221024.0 & $6.51e-100$ \\
        Coll. Freq.     & 128256.0 & $0.312171$ & 137374.0 & $0.000310$ \\
        Frozen Freq.    & 73315.5  & $2.10e-44$ & 127930.0 & $0.323385$ \\
        \hline
        \multicolumn{5}{|c|}{\rule{0pt}{2ex} {SICNav-p vs. RGL}} \\
        \hline
        Nav. Time       & 221622.0 & $1.76e-100$& 37619.5  & $7.47e-88$ \\
        Coll. Freq.     & 135752.5 & $0.000300$ & 144991.0 & $4.08e-09$ \\
        Frozen Freq.    & 120759.0 & $0.112056$ & 20241.5  & $2.21e-130$ \\
        \hline
        \end{tabular}
    \end{center}
\end{table}

Tables~\ref{tab:pairwise_sicnav_vs_baselines}~and~\ref{tab:pairwise_sicnav_vs_baselines_sfm} summarize the results of the Mann-Whitney U test for the comparison of SICNav-p against each of the baseline algorithms for the ORCA simulation and SFM simulation, respectively. We observe that almost all the p-values for the Mann-Whitney U test are low for all metrics and navigation scenarios, indicating that there are significant differences between SICNav-p and each of the baseline algorithms. The only exceptions are the frozen frequency metric in the $N = 5$ humans environment for the comparison of SICNav-p against SARL in both the ORCA and SFM simulations, where the p-value exceeds $\alpha=0.05$. This indicates that there are no significant differences between SICNav-p and SARL for the frozen frequency metric in the $N = 5$ humans environment in these simulations.
\begin{table}[t]
    \color{blue}
    \caption{\color{blue}Pairwise Mann-Whitney U Test Results of SICNav-np against Baseline Algorithms in \ac{ORCA} simulations. Arrows in the table ($\uparrow$ for Stat. and $\downarrow$ for p-value) represent our analytical desire to uncover significant differences between SICNav-np and each baseline, indicating SICNav-np's distinct performance advantages.}
    \label{tab:pairwise_sicnav_np_vs_baselines}
    \begin{center}
        \begin{tabular}{|l|c|c|c|c|}
        \hline
        & \multicolumn{2}{c|}{$N = 3$ humans} & \multicolumn{2}{c|}{$N = 5$ humans} \\
        \cline{2-5}
        \textbf{Metric} & \textbf{Stat.} $\uparrow$ & \textbf{p-value} $\downarrow$ & \textbf{Stat.} $\uparrow$ & \textbf{p-value} $\downarrow$ \\
        \hline\hline
         \multicolumn{5}{|c|}{\rule{0pt}{2ex} {SICNav-np vs. MPC-CVMM}} \\
        \hline
        Nav. Time       & 89875.0 & $7.78e-15$ & 112744.5 & $0.007183724$ \\
        Coll. Freq. & 115875.0 & $0.000162293$ & 97726.0 & $1.62e-13$ \\
        Frozen Freq.    & 111069.0 & $6.80e-07$ & 104256.5 & $2.08e-09$ \\
        \hline
         \multicolumn{5}{|c|}{\rule{0pt}{2ex} {SICNav-np vs. ORCA}} \\
        \hline
        Nav. Time       & 41541.5 & $1.55e-74$ & 25943.5 & $5.60e-104$ \\
        Coll. Freq. & 117143.5 & $0.001753694$ & 116393.0 & $0.018405429$ \\
        Frozen Freq.    & 84537.0 & $7.19e-31$ & 79896.5 & $1.10e-30$ \\
        \hline
        \multicolumn{5}{|c|}{\rule{0pt}{2ex} {SICNav-np vs. DWA}} \\
        \hline
        Nav. Time       & 5706.5 & $3.43e-152$ & 10793.5 & $6.73e-138$ \\
        Coll. Freq. & 120028.0 & $0.035176261$ & 88519.0 & $1.51e-19$ \\
        Frozen Freq.    & 24003.5 & $1.42e-127$ & 18627.0 & $5.13e-132$ \\
        \hline
        \multicolumn{5}{|c|}{\rule{0pt}{2ex} {SICNav-np vs. SARL}} \\
        \hline
        Nav. Time       & 164838.5 & $1.19e-18$ & 216830.0 & $9.26e-91$ \\
        Coll. Freq. & 113773.0 & $7.73e-06$ & 105817.0 & $6.84e-08$ \\
        Frozen Freq.    & 95694.5 & $2.90e-20$ & 130540.0 & $0.053446485$ \\
        \hline
        \multicolumn{5}{|c|}{\rule{0pt}{2ex} {SICNav-np vs. RGL}} \\
        \hline
        Nav. Time       & 180210.5 & $3.35e-34$ & 108649.0 & $0.000304544$ \\
        Coll. Freq. & 113542.0 & $4.59e-06$ & 94052.5 & $2.32e-15$ \\
        Frozen Freq.    & 117233.5 & $0.002776$ & 67023.0 & $4.91e-51$ \\
        \hline
        \end{tabular}
    \end{center}
\end{table}

\begin{table}[t]
    \color{blue}
    \caption{\color{blue}Pairwise Mann-Whitney U Test Results of SICNav-np against Baseline Algorithms in \ac{SFM} simulations. Arrows in the table ($\uparrow$ for Stat. and $\downarrow$ for p-value) represent our analytical desire to uncover significant differences between SICNav-np and each baseline, indicating SICNav-np's distinct performance advantages.}
    \label{tab:pairwise_sicnav_np_vs_baselines_sfm}
    \begin{center}
        \begin{tabular}{|l|c|c|c|c|}
        \hline
        & \multicolumn{2}{c|}{$N = 3$ humans} & \multicolumn{2}{c|}{$N = 5$ humans} \\
        \cline{2-5}
        \textbf{Metric} & \textbf{Stat.} $\uparrow$ & \textbf{p-value} $\downarrow$ & \textbf{Stat.} $\uparrow$ & \textbf{p-value} $\downarrow$ \\
        \hline\hline
         \multicolumn{5}{|c|}{\rule{0pt}{2ex} {SICNav-np vs. MPC-CVMM}} \\
        \hline
        Nav. Time       & 105022.0 & $1.55e-08$ & 128497.0 & $0.223828$ \\
        Coll. Freq.     & 130855.5 & $0.381082$ & 125318.0 & $0.121516$ \\
        Frozen Freq.    & 119577.0 & $1.40e-05$ & 116662.5 & $0.014878$ \\
        \hline
         \multicolumn{5}{|c|}{\rule{0pt}{2ex} {SICNav-np vs. ORCA}} \\
        \hline
        Nav. Time       & 57897.0  & $2.42e-38$ & 66466.5  & $2.15e-37$ \\
        Coll. Freq.     & 108882.0 & $4.65e-09$ & 141355.5 & $0.274905$ \\
        Frozen Freq.    & 144661.0 & $0.960146$ & 127993.5 & $0.011287$ \\
        \hline
        \multicolumn{5}{|c|}{\rule{0pt}{2ex} {SICNav-np vs. DWA}} \\
        \hline
        Nav. Time       & 13089.0  & $1.79e-104$ & 12064.0  & $6.03e-89$ \\
        Coll. Freq.     & 137814.0 & $0.724094$ & 122684.0 & $0.015292$ \\
        Frozen Freq.    & 84822.0  & $1.63e-80$ & 79184.0  & $8.21e-63$ \\
        \hline
        \multicolumn{5}{|c|}{\rule{0pt}{2ex} {SICNav-np vs. SARL}} \\
        \hline
        Nav. Time       & 164449.5 & $5.09e-25$ & 212318.0 & $2.07e-66$ \\
        Coll. Freq.     & 127842.5 & $0.148963$ & 112235.5 & $3.13e-05$ \\
        Frozen Freq.    & 128716.5 & $0.235351$ & 129313.5 & $0.073007$ \\
        \hline
        \multicolumn{5}{|c|}{\rule{0pt}{2ex} {SICNav-np vs. RGL}} \\
        \hline
        Nav. Time       & 198035.0 & $1.45e-46$ & 101080.0 & $9.73e-06$ \\
        Coll. Freq.     & 126678.0 & $0.107036$ & 146464.0 & $0.946688$ \\
        Frozen Freq.    & 107908.5 & $1.78e-10$ & 88899.0  & $8.11e-63$ \\
        \hline
        \end{tabular}
    \end{center}
\end{table}

Tables~\ref{tab:pairwise_sicnav_np_vs_baselines}~and~\ref{tab:pairwise_sicnav_np_vs_baselines_sfm} summarize the results of the Mann-Whitney U test for the comparison of SICNav-np against each of the baseline algorithms for the ORCA simulation and SFM simulation, respectively. We can make the same observations as for SICNav-p, where almost all the p-values for the Mann-Whitney U test are low for all metrics and navigation scenarios, indicating that there are significant differences between SICNav-np and each of the baseline algorithms. The only exceptions are the frozen frequency metric in the $N = 5$ humans environment in the ORCA simulation and the collision frequency metric in the $N = 3$ humans environment in the SFM simulation for the comparison of SICNav-np against SARL, where the p-value exceeds $\alpha=0.05$. This indicates that there are no significant differences between SICNav-np and SARL for these metrics and environments.

\subsubsection{Discussion and Conclusion}
In this section we performed a series of statistical tests to verify the significance of the crowd navigation performance results that we present in Section~\ref{sec:sim_results}.
In the first series of tests, we compared the two variants of our algorithm, SICNav-p and SICNav-np, with each other. We found that the absence of privileged information in SICNav-np does not significantly affect most metrics. This finding led us to compare each variant of our method with each of the baseline algorithms in pairwise fashion. Combining our findings from these pairwise statistical comparisons with the average values of metrics presented in Section~\ref{sec:sim_results}, we show that for all metrics and navigation scenarios both variants of our method significantly outperformed all the baselines. The only exception was the frozen frequency metric in the $N = 5$ humans environment for the comparisons of the variants of our method with SARL.
Our analysis provides compelling evidence that the proposed method is significantly better than the baseline algorithms for crowd navigation in both the $N = 3$ humans and $N = 5$ humans environments for both \ac{ORCA} and \ac{SFM} simulators.
}
\section*{Acknowledgement}
We acknowledge the support of the Natural Sciences and Engineering Research Council of Canada (NSERC).

Computing resources used in preparing this research were provided, in part, by the Province of Ontario, the Government of Canada through CIFAR, and companies sponsoring the Vector Institute for Artificial Intelligence. %

\bibliographystyle{IEEEtran}
\bibliography{IEEEabrv, references}

\begin{IEEEbiography}[{\includegraphics[width=1in,height=1.25in,clip,keepaspectratio]{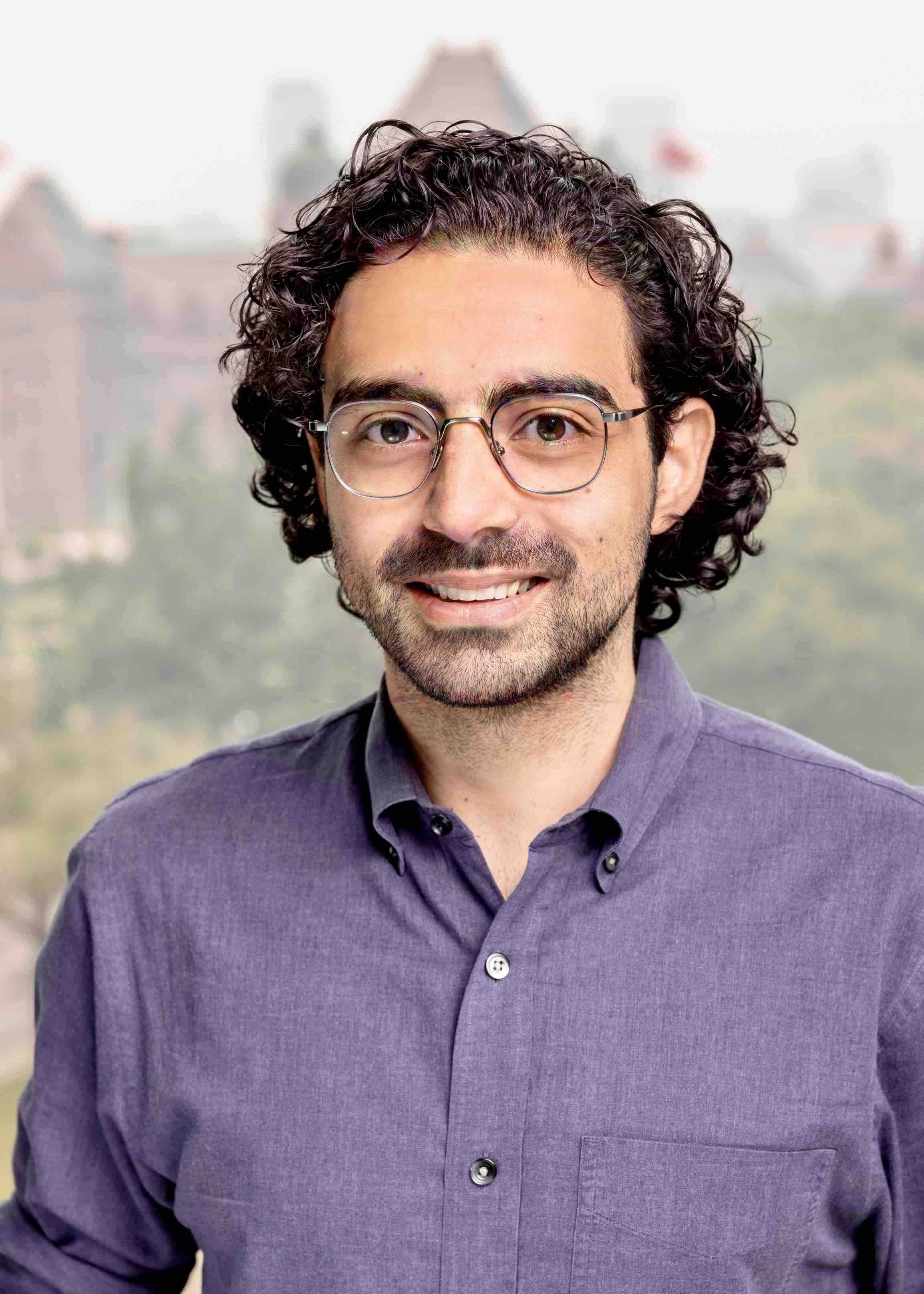}}]{Sepehr Samavi} is a Ph.D. Candidate at the University of Toronto Institute for Aerospace Studies, Canada. He received the B.A.Sc. in Engineering Science and M.A.Sc from the University of Toronto. His research interests lie at the intersection of control, optimization, and machine learning for robotics, with the goal to empower safe navigation in unstructured, dynamic, and human-centered environments. His research has been supported by the Alexander Graham Bell Canada Graduate Scholarship (CGSD) by the Natural Sciences and Engineering Research Council of Canada (NSERC).
\end{IEEEbiography}

{\color{blue}
\begin{IEEEbiography}[{\includegraphics[width=1in,height=1.25in,clip,keepaspectratio]{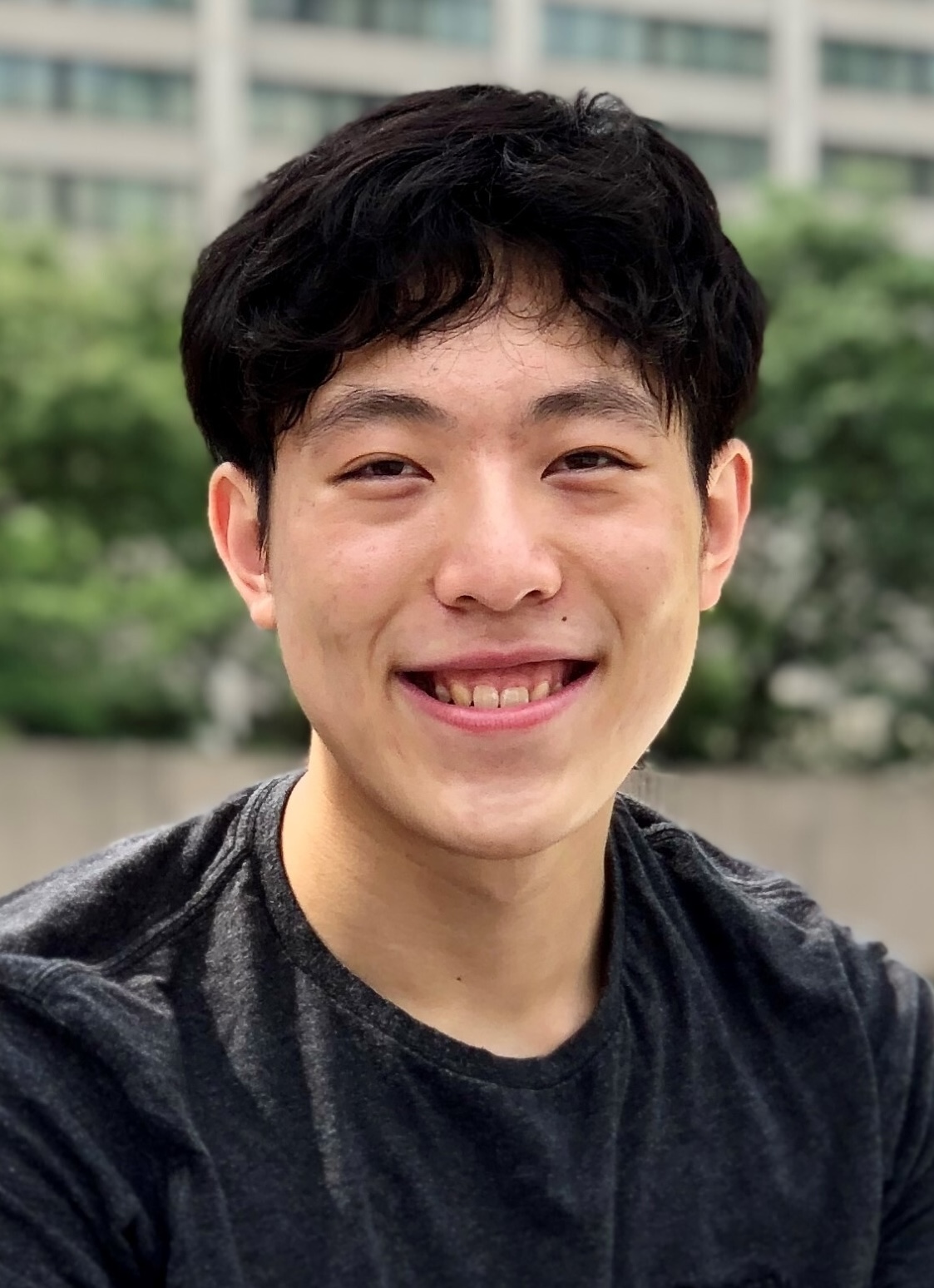}}]{James R. Han} is a M.A.Sc student at the University of Toronto Institute for Aerospace Studies, Canada. He earned his B.A.Sc. in Engineering Science with a focus on Machine Intelligence and Robotics from the University of Toronto. His academic pursuits are centered at the intersection of Reinforcement Learning, Deep Learning, and Robotics, aiming to bring about real-world robots. This work was conducted during his undergraduate studies at the University of Toronto.
\end{IEEEbiography}
}
\begin{IEEEbiography}[{\includegraphics[width=1in,height=1.25in,clip,keepaspectratio]{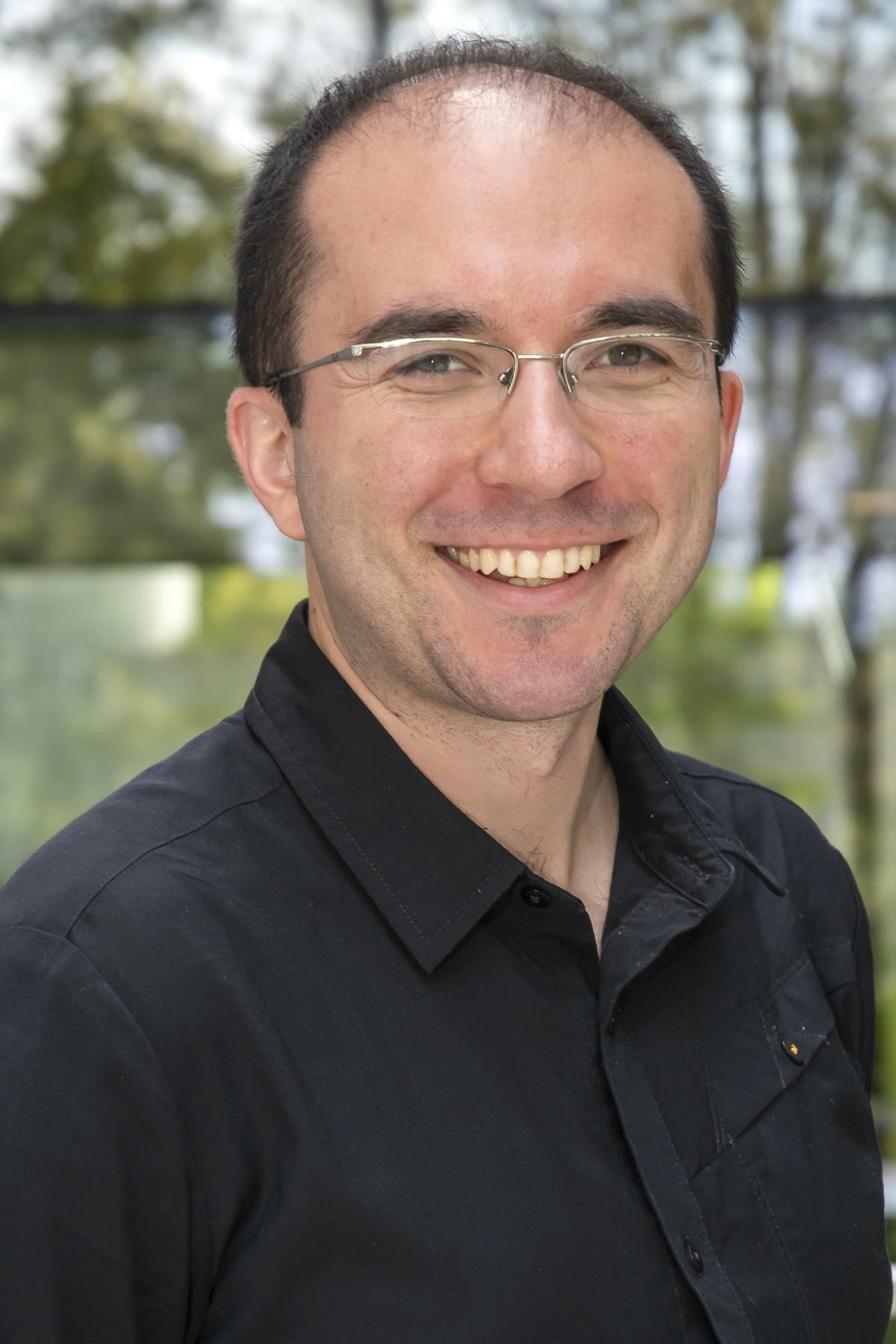}}]{Florian Shkurti} is an Assistant Professor in the Robotics Group at the Department of Computer Science, University of Toronto, Canada. He directs the Robot Vision and Learning (RVL) lab. His research centers around robotics and spans machine learning, perception, planning and control, with the aim of developing methods that enable robots to perceive, reason, and act effectively and safely, particularly in dynamic environments and alongside humans. He received his PhD from McGill University.
\end{IEEEbiography}

\begin{IEEEbiography}[{\includegraphics[width=1in,height=1.25in,clip,keepaspectratio]{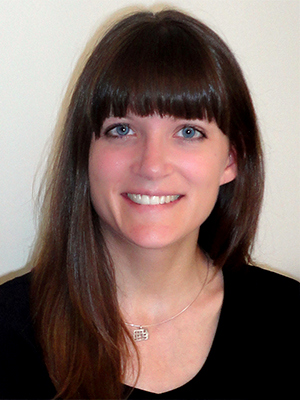}}]{Angela P. Schoellig} is an Alexander von Humboldt Professor for Robotics and Artificial Intelligence at the Technical University of Munich, Germany. She is also an Associate Professor at the University of Toronto Institute for Aerospace Studies and a Faculty Member of the Vector Institute in Toronto, Canada. Angela conducts research at the intersection of robotics, controls, and machine learning with the goal of enhancing the performance, safety, and autonomy of robots by enabling them to learn from experience. She received her Dr. sc. (Ph.D.) from ETH Zurich in 2013, where she was awarded the ETH Medal and the Dimitris N. Chorafas Foundation Award.
\end{IEEEbiography}

\end{document}